\documentclass[10pt,twocolumn,letterpaper]{article}

\usepackage{iccv}

\usepackage{times}
\usepackage{epsfig}
\usepackage{graphicx}
\usepackage{amsmath}
\usepackage{amssymb}
\usepackage{booktabs} %
\usepackage{enumitem}
\usepackage{diagbox}
\usepackage{tabularx}
\usepackage{multirow}
\usepackage{booktabs}
\usepackage[table,xcdraw]{xcolor}
\usepackage[ruled,vlined]{algorithm2e}
\usepackage{float}
\usepackage{caption}
\usepackage{subcaption}
\usepackage[numbers,sort,compress]{natbib}
\usepackage{esvect}
\usepackage{dblfloatfix}%

\SetAlFnt{\small}

\usepackage[pagebackref=true,breaklinks=true,letterpaper=true,colorlinks,bookmarks=false]{hyperref}

\iccvfinalcopy %

\def\iccvPaperID{8024} %

\newcommand{\mpc}{MPC\xspace}

\newcommand{\labelSet}{L}
\newcommand{\labelSingle}{l}

\newcommand{\labelSetVertexVector}{\mathbf{v}}
\newcommand{\labelSetVariedVertexVector}{\tilde{\labelSetVertexVector}}

\newcommand{\numMarkers}{M}

\newcommand{\pointSet}{P}

\newcommand{\asmat}{A}
\newcommand{\augasmat}{\asmat^{'}}

\newcommand{\asmatGt}{G}
\newcommand{\augasmatGt}{\asmatGt^{'}}

\newcommand{\asmatPred}{\asmat}
\newcommand{\augasmatPred}{\augasmat}

\newcommand{\ascore}{S}
\newcommand{\ascorePred}{\ascore}
\newcommand{\joints}{J}

\newcommand{\markers}{X}

\newcommand{\bodyVertex}{V}
\newcommand{\markerDistance}{d}

\newcommand{\lossAsmatWeight}{W}

\newcommand{\loss}{\mathbb{L}}

\newcommand{\numAttHeads}{h}
\newcommand{\numAttLayers}{k}

\newcommand{\mocap}{mocap\xspace}
\newcommand{\Mocap}{MoCap\xspace}

\newcommand{\soma}{SOMA\xspace}

\newcommand{\amass}{AMASS\xspace}

\newcommand{\vicon}{{Vicon}\xspace}
\newcommand{\shogun}{\xspace{Sh\={o}gun}\xspace}

\newcommand{\phasespace}{{PhaseSpace}\xspace}

\newcommand{\cmuii}{{CMU-II}\xspace}

\newcommand{\validationDataset}{{HDM05}\xspace}
\newcommand{\citeValidationDataset}{\cite{MPI_HDM05}\xspace}
\newcommand{\valds}{\validationDataset}

\newcommand{\mosh}{MoSh\xspace}
\newcommand{\moshpp}{\mosh}

\newcommand{\citemosh}{\cite{Loper:SIGASIA:2014,AMASS:2019}\xspace}

\newcommand{\smplx}{SMPL-X\xspace}

\newcommand{\scbodymodelcite}{\smplx \cite{SMPL-X:2019}\xspace}

\newcommand{\nan}{null\xspace}

\newcommand{\amassHours}{45 hours\xspace}
\newcommand{\numSmplxBetasNeutral}{3664\xspace}
\newcommand{\unifiedFramRate}{30 Hz\xspace}
\newcommand{\numHoursNewMoCapProcessed}{8\xspace}
\newcommand{\numNewDatasetsProcessed}{4 \xspace}
\newcommand{\numSinkhornIter}{35\xspace}
\newcommand{\numSOMAParameters}{$1.44$ M\xspace}
\newcommand{\trainTime}{3\xspace}

\newcommand{\learningRate}{$1e-3$\xspace}
\newcommand{\weightDecay}{$5e-5$\xspace}
\newcommand{\numAttentionLayers}{$8$\xspace}
\newcommand{\earlyStopPatience}{$8$\xspace}
\newcommand{\lrSchedulerPatience}{$3$\xspace}
\newcommand{\numMarkersSuperSet}{$89$\xspace}

\newcommand{\gt}{ground truth\xspace}

\usepackage[numbers,sort]{natbib}

\begin{document}

\newcommand*\samethanks[1][\value{footnote}]{\footnotemark[#1]}
\newcommand*{\affaddr}[1]{#1} 
\newcommand*{\affmark}[1][*]{\textsuperscript{#1}}
\newcommand*{\email}[1]{\small{\texttt{#1}}}

\title{\soma: Solving Optical Marker-Based MoCap Automatically}

\author{Nima Ghorbani \hspace{0.3in} 
	Michael J. Black
	\\
	{Max Planck Institute for Intelligent Systems, T\"ubingen, Germany} \\
	\email{\{nghorbani,black\}@tuebingen.mpg.de}
}

\twocolumn[{%
	\renewcommand\twocolumn[1][]{#1}%
	\vspace*{-2.0em}
	\maketitle
	\begin{center}
		\vspace*{-0.2in}
		\centerline{
			\includegraphics[width=1\linewidth]{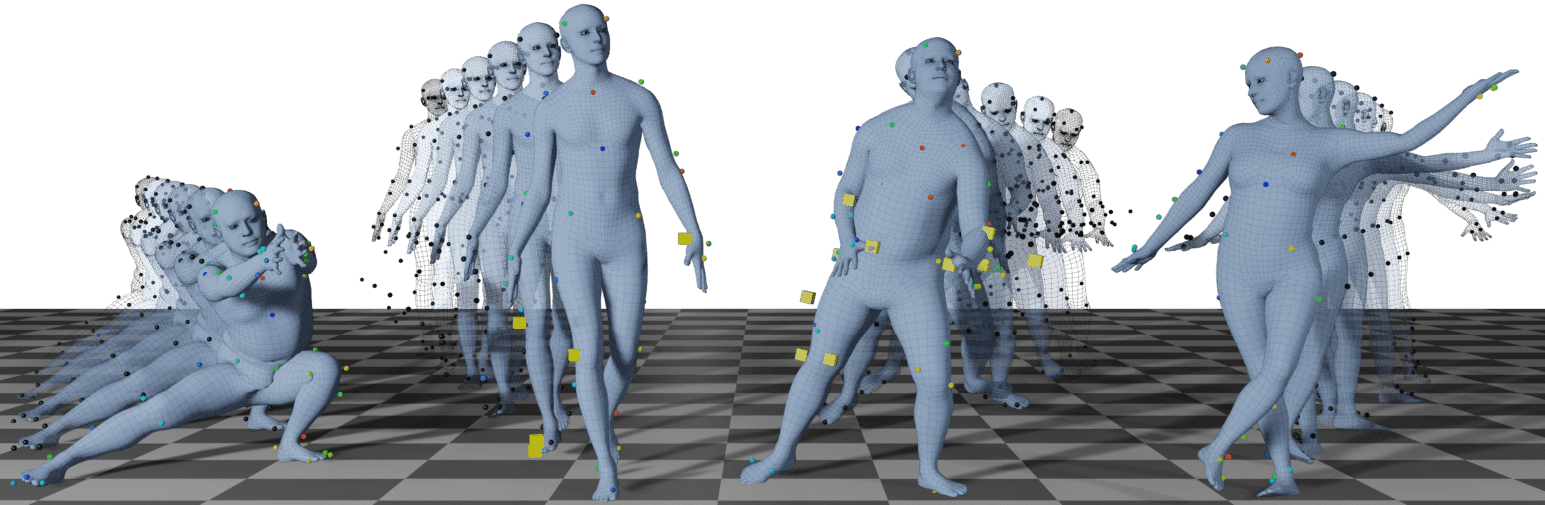}
		}
				\vspace{-0.5em}
		\captionof{figure}{
			\soma transforms raw motion capture (\mocap) point clouds
			(black dots) to labeled markers (colored dots). 
			The cubes in yellow are detected ``ghost''
			points; e.g.~spurious reflections, non-body markers, or unidentified points. 
			With labeled \mocap we fit \smplx bodies (blue
			mesh) using
			\mosh~\citemosh. %
		}
		\label{fig:teaser}
		\vspace{-0.5em}
	\end{center}%
}]

\maketitle
\ificcvfinal\thispagestyle{empty}\fi

\begin{abstract}\label{sec:abstract}
\vspace*{-1.6em}

Marker-based optical motion capture (\mocap) is the ``gold standard'' method for acquiring accurate 3D human motion in  computer vision, medicine, and graphics. 
The raw output of these systems are noisy and incomplete 3D points or short tracklets of points.
To be useful, one must associate these points with corresponding markers on the captured subject; i.e.~``labelling''.
Given these labels, one can then ``solve'' for the 3D skeleton or body surface mesh.
Commercial auto-labeling tools require a specific calibration procedure at capture time,  which is not possible for archival data.
Here we train a novel neural network called \soma, which takes raw \mocap point clouds with varying numbers of points, labels them at scale without any calibration data, independent of the capture technology, and requiring only minimal human intervention.
Our key insight is that, while labeling point clouds is highly ambiguous, the 3D body provides strong constraints on the solution that can be exploited by a learning-based method. %
To enable learning, we generate massive training sets of simulated noisy and \gt \mocap markers animated by 3D bodies from \amass. 
\soma exploits an architecture with stacked 
self-attention elements to learn the spatial structure of the 3D body and an optimal transport layer %
to constrain the assignment (labeling) problem while rejecting outliers. 
We extensively evaluate \soma both quantitatively and qualitatively.
\soma is more accurate and robust than existing state of the art research methods and can be applied where commercial systems cannot.
We automatically label over \numHoursNewMoCapProcessed hours of archival \mocap data across \numNewDatasetsProcessed different datasets captured using various technologies and output \smplx body models.
The model and data is released for research purposes at \url{https://soma.is.tue.mpg.de/}. %
\end{abstract}

\section{Introduction}\label{sec:introduction}

Marker-based optical motion capture (\mocap) systems record 2D infrared images of light reflected or emitted by a set of markers placed at key locations on the surface of a subject's body. %
Subsequently, the \mocap systems recover the precise position of the markers as a sequence of sparse and unordered points or short tracklets. %
Powered by years of commercial development, these systems offer high temporal and spatial accuracy. %
Richly varied \mocap data from such systems is widely used to train machine learning methods in action recognition, motion synthesis, human motion modeling, pose estimation, etc.
Despite this, the largest existing \mocap dataset, \amass \cite{AMASS:2019}, has about \amassHours of \mocap, much smaller than video datasets used in the field.

Mocap data is limited since capturing and  processing it is expensive. 
Despite its value, there are large amounts of archival \mocap in the world that have never been labeled; this is the ``dark matter'' of \mocap.
The problem is that, to solve for the 3D body, the raw \mocap point cloud (\mpc) must be ``labeled''; that is, the points must be assigned to physical ``marker'' locations on the subject's body.
This is challenging because the \mpc is noisy and sparse and the labeling problem is ambiguous.
Existing commercial tools, e.g.\ \cite{2019optitrack,2019vicon}, offer partial automation,
however none provide a complete solution 
to automatically handle variations in marker layout, i.e.\ number of markers used and their rough placement on the body, variation in subject body shape and gender, and variation across capture technologies namely active vs passive markers or brands of system.
These challenges typically preclude cost-effective labeling of archival data, and add to the cost of new captures by requiring manual cleanup.

Automating the \mocap labeling problem has been examined by the research community \cite{ghorbanietemadetal2019, hanliuetal2018,Holden:2018:Robust}. %
Existing methods focus on fixing the mistakes in already labeled markers through denoising \cite{Holden:2018:Robust, mocapsolver2021}.
Recent work formulates the problem in a matching framework, %
directly predicting the label assignment matrix for a fixed number of markers in a restricted setup \cite{ghorbanietemadetal2019}. 
In short, the existing methods are limited to a constrained range of motions \cite{ghorbanietemadetal2019}, a single body shape \cite{hanliuetal2018,Holden:2018:Robust, mocapsolver2021}, a certain capture scenario, a special marker layout, or require a subject-specific calibration sequence \cite{ghorbanietemadetal2019, 2019optitrack,2019vicon}. %
Other methods require high-quality real \mocap marker data for training, effectively prohibiting their scalability to novel scenarios \cite{mocapsolver2021, ghorbanietemadetal2019}.

To address these shortcomings we take a data-driven approach and train a neural network end-to-end with self-attention components and an optimal transport layer to predict a per-frame constrained inexact matching between \mocap points and labels.
Having enough ``real'' data %
for training is not feasible, %
therefore we opt for synthetic data. Given a marker layout, we generate synthetic \mocap point clouds %
with realistic noise, and then train a layout-specific network that can cope with realistic variations across a whole \mocap dataset. %
While previous works have exploited synthetic data \cite{Holden:2018:Robust, hanliuetal2018}, they are limited in terms of body shapes, motions, marker layouts, and noise sources.

Even with a large synthetic corpus of \mpc, labeling a cloud of sparse 3D points, containing outliers and missing data, is a highly ambiguous task.
The key to the solution lies in the fact that the points are structured, as is their variation with articulated pose. %
Specifically, they are constrained by the shape and  motion of the human body.
Given sufficient training data, our attentional framework learns to exploit local context at different scales.
Furthermore, %
if there were no noise, the mapping between labels and points would be one-to-one.
We formulate these concepts as a unified training objective enabling end-to-end model training.
Specifically, our formulation exploits a transformer architecture to capture local and global contextual information using self-attention (Sec.~\ref{sec:method_self_attention}).
By generating synthetic \mocap data with varying body shapes and poses, \soma implicitly learns %
the kinematic constraints of the underlying deformable human body (Sec.~\ref{sec:method_synthetic_data}). 
A one-to-one match between 3D points and markers, subject to missing and spurious data, is achieved by a special normalization technique %
(Sec.~\ref{sec:method_constrained_point_labeling}).
To provide a common output framework, consistent with \cite{AMASS:2019}, we use \moshpp \citemosh as a post-processing step to fit \scbodymodelcite to the labeled points; this also helps deal with missing data caused by occlusion or dropped markers.
The \soma system is outlined in Fig.~\ref{fig:sc_architecture}.

To generate training data, \soma requires a rough marker layout that can be obtained by a single labeled frame, which requires minimal effort. 
Afterward, virtual markers are automatically placed on a \smplx body and animated by motions from AMASS \cite{AMASS:2019}. 
In addition to common \mocap noise models like occlusions \cite{ghorbanietemadetal2019,hanliuetal2018,Holden:2018:Robust}, and ghost points \cite{hanliuetal2018,Holden:2018:Robust}, we introduce novel terms to vary maker placement on the body surface and we copy noise from real marker data in \amass (Sec.~\ref{sec:method_synthetic_data}). %
We train \soma once for each \mocap dataset and apart from the one layout frame, we do not require any labeled real data. %
After training, given a noisy \mpc \emph{frame} as input, \soma predicts a distribution over labels of each point, including a \nan label for ghost points. %

We evaluate \soma on several challenging datasets and find that we outperform the current state of the art numerically while being much more general.
Additionally, we capture new \mpc data using a %
\vicon \mocap system and compare hand-labeled ground-truth to \shogun and \soma output.
\soma performs similarly compared with the commercial system.
Finally, we apply the method on archival \mocap datasets:
Mixamo \cite{mixamoWEB},
DanceDB \cite{DanceDB:2018:aristidou}, and a previously unreleased portion of the CMU \mocap dataset \cite{cmuWEB}. 

In summary, our main contributions are:
(1) a novel neural network architecture exploiting self-attention to process sparse deformable point cloud data;
(2) a system that consumes \mocap point clouds directly and outputs a distribution over marker labels; %
(3) a novel synthetic \mocap generation pipeline that generalizes to real \mocap datasets;
(4) a robust solution that works with archival data, different \mocap technologies, poor data quality, and varying subjects and motions;
(5) 220 minutes of processed \mocap data in \smplx format, trained models, and code are released for research purposes.

\section{Related Work}\label{sec:relatedwork}

\textbf{Learning to Process \Mocap Data} was first introduced by \cite{Song2003} in a limited scenario.
More recently, %
\cite{ghorbanietemadetal2019} proposes a learning-based model for \mocap labeling that directly predicts %
permutations of 44 input markers. 
The number of possible permutations is prohibitive, %
hence, the authors restrict them %
to a limited pool shown to the network during training and test time. %
Moreover, motions are restricted to four categories:  walk, jog, jump and sit. 
Furthermore, \cite{ghorbanietemadetal2019} inherently cannot deal with ghost points.
We compare directly with them and find that we are more accurate, while removing the limitations.

Solutions exist that ``denoise'' the possible incorrect labels of already labeled \mocap markers \cite{Holden:2018:Robust, mocapsolver2021}. 
These approaches %
normalize the markers to a standard body size, and rely on fragile heuristics to remove ghost points, and must first compute the global orientation of the body. %
Our method starts one step earlier, with {\em unlabeled point clouds}, and outputs a fully labeled sequence, while learning to reject ghost points and dealing with varied body shapes. %

\textbf{Deep Learning on Point Cloud Data}
requires a solution to handle a variable number of unordered points, and a way to define a notion of ``neighborhood''. 
To address these issues \cite{ge_cvpr16_3d,su15mvcnn} project 3D points into 2D images and \cite{voxnet_iros2015_maturana, zhirongwusongetal2015b}  rasterize the point cloud into 3D voxels to enable the use of traditional convolution operators.
Han et al.~\cite{hanliuetal2018} utilize this idea for labeling hand \mpc data, by projecting them into multiple 2D depth images.
Using these, they train a neural network to predict the 3D location of 19 known markers on a fixed-shape hand, assigning the label of a marker to a point closest to it in a disconnected graph matching step.
In contrast, our pipeline directly works with \mocap point clouds and predicts a distribution over labels for each point, end-to-end, without disconnected stages. %

PointNet methods \cite{pointnet_charles_2017,pointnetpp_2017} also process the 3D point cloud directly, while  learning local features with permutation-invariant pooling operators. 
Further non-local networks \cite{NonLocal2018} and self-attention-based \cite{vaswaniAttentionAllYou2017} models can attend globally while learning to focus locally on specific regions of the input. 
This simple formulation enables learning robust features on  sparse point clouds while being insensitive to variable numbers of points. 
\soma is a novel application of this idea to \mocap data.
In Sec.~\ref{sec:method_self_attention} we demonstrate that by stacking multiple self-attention elements, \soma can learn rich point features at multiple scales that enable robust,  permutation-invariant, \mocap labeling.

\textbf{Inexact Graph Matching} formulates the problem of finding an assignment between nodes of a model
graph to data nodes, where the former occasionally has fewer elements than the latter. %
This is an NP-hard problem \cite{Abdulkader_1998} in general and challenging for \mocap due to occlusions and spurious data points.
Such graph matching problems appear frequently in computer vision and are addressed either with engineered costs \cite{berg_2005_shape_matching,Torresani_2008, Leordeanu_2005} or learned ones \cite{Caetano_2007_graph_matching,SuperGlue_Sarlin_2020_CVPR}. 
Ghorbani et al.~\cite{ghorbanietemadetal2019} apply this to \mocap using an approximate solution with Sinkhorn normalization \cite{adams2011ranking,sinkhorn1967} by assuming that the graph of a \mocap frame is isomorphic to the graph of the labels. 
We relax this assumption by considering an \emph{inexact match} between the labels and points, %
by opting for an \emph{optimal transport} \cite{cedric_2008_optimal_transport} solution.

\textbf{Body Models}, in the form of skeletons,   %
are widely used to constrain the labeling problem and to solve for body pose \cite{herdafuaetal2001, Ringer2002,ringerlasenby2002, meyerkudereretal2014, contini1972, schubertgkogkidisetal2015, schubertgkogkidisetal2015}.
Recently \cite{Steinbring2016, joukovetal2020} employ variants of Kalman filtering to estimate constrained configuration of a given body skeleton in real-time but are susceptible to occlusions and ghost points.
The most mature commercial solution to-date is \shogun \cite{2019vicon}, which can produce real-time labeling and skeletal solving.  
While this is an excellent product, out-of-the-box it works only with a \vicon-specific marker layout and requires subject-specific session calibration.  
Thus it is not a general solution and cannot be used on most archival \mocap data or with customized marker layouts needed in many applications.
\moshpp \citemosh, goes beyond simple skeleton models and utilizes a realistic 3D body surface model learned from a large corpus of body scans \cite{SMPL-X:2019, GHUM2020, FRANK2018}. %
It takes  labeled \mocap markers as well as a rough layout of their placement on the body, and solves for body parameters and the precise placement of markers on the body. 
We employ \moshpp for post-processing auto-labeled \mocap across different datasets into a unified \smplx representation.

\section{The \Mocap Labeling Problem}\label{sec:what_is_mocap}
\begin{figure}[tbh] 
	\centering
	\includegraphics[width=.9\linewidth]{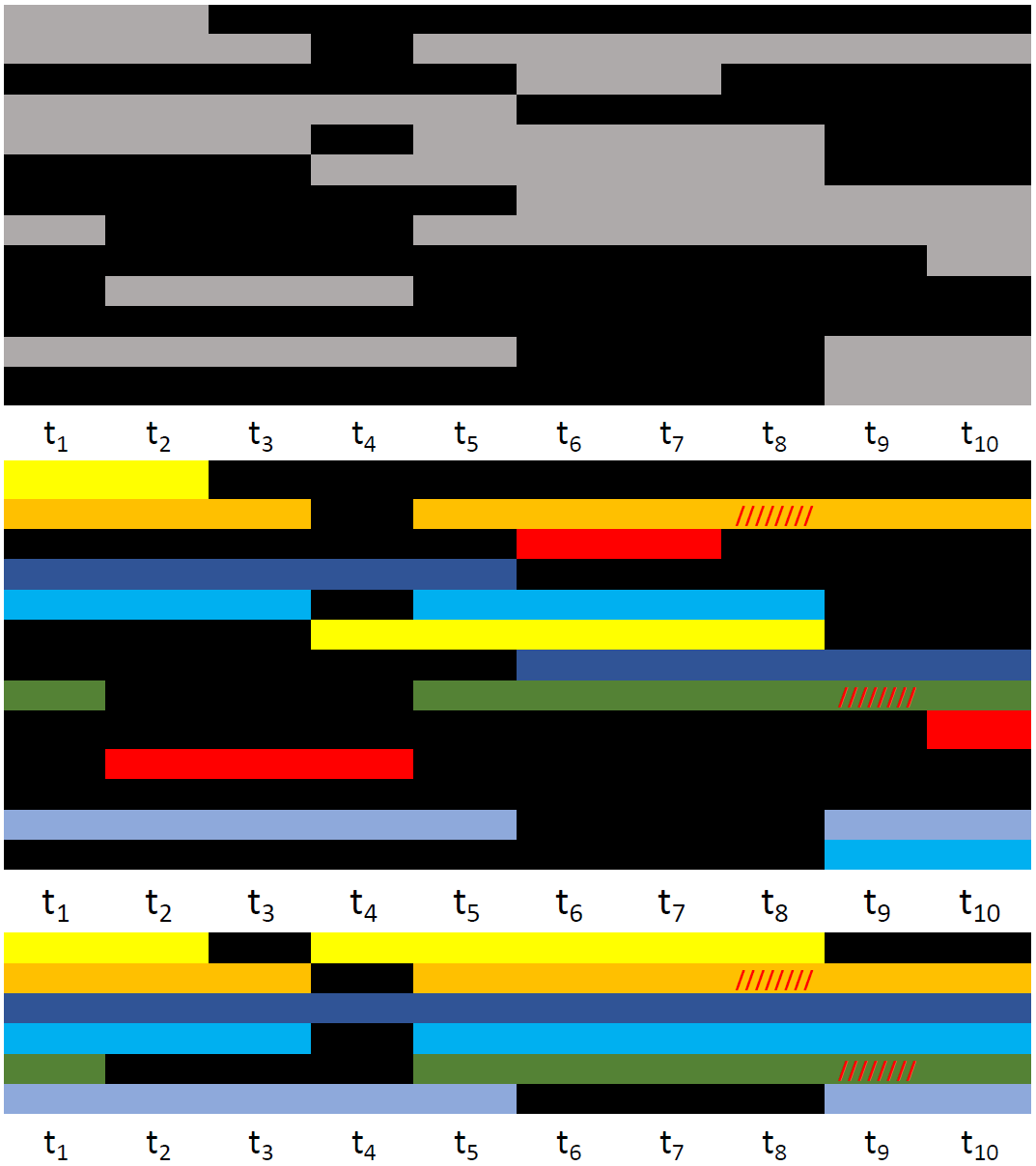}
	\caption{\Mocap labeling problem. 
		 (top) Raw, unlabeled, \Mocap Point Cloud (\mpc).
                 Each column represents a timestamp in a
                \mocap sequence and each cell is a 3D point or
                a short tracklet (shown as a gray row).  
		(middle) shows the \mpc after labeling. Colors correspond
                to different labels in the marker layout. Red
                corresponds to ghost points (outliers).
                The red oblique lines show ghost points wrongly
                tracked as actual markers by the \mocap system. 
		 (bottom) shows the final result, with the tracklets glued
                 together to form full trajectories with only valid
                 markers retained. 
                 Note that marker occlusion results in black (missing) sections.
	}
	\label{fig:what_is_mocap}
	\vspace*{-1.5em}
\end{figure}

A \mocap point cloud, \mpc, is a time sequence with $T$ frames of 3D points
\begin{eqnarray}\label{eqn:mpc}
	\mathrm{\mpc} & = &  \{ \pointSet_1, \ldots, \pointSet_T \},\\
	\pointSet_{t} & = & \{ \pointSet_{t,1}, \ldots, \pointSet_{t,n_t} \},
  \mbox{\hspace{0.1in}} \pointSet_{t,i} \in \mathbb{R}^3,
\end{eqnarray}
where $|\pointSet_{t}| = n_t$ for each time step $t \in \{1:T\}$. 
We visualize an \mpc as a chart in Fig.~\ref{fig:what_is_mocap} (top), where each row holds points reconstructed by the \mocap hardware, and each column represents a frame of \mpc.
Each point is unlabeled but these can often be locally tracked over short time intervals, illustrated by the gray bars in the figure.
Note that some of these tracklets may correspond to noise or ``ghost points''.
For passive marker systems like \vicon \cite{2019vicon}, a point that
is occluded typically appears in a new row; i.e.\ with a new ID. 
For active marker systems like \phasespace \cite{2019phasespace}, one can have gaps in the trajectories due to occlusion.
The figure shows both types.

The goal of \mocap labeling is to assign each point (or tracklet) to a
corresponding marker label 
\begin{equation}
	\labelSet = \{\labelSingle_1, \dots, \labelSingle_\numMarkers, \mathrm{\nan}\},
\end{equation}
in the marker layout as illustrated in Fig.~\ref{fig:what_is_mocap} (middle), where each color is a different label.
The set of marker labels include an extra \nan label for points that are not valid markers, hence $|\labelSet| = \numMarkers + 1$.  
These are shown as red in the figure.
Valid point labels and tracklets of them are subject to several constraints: 
\phantomsection\label{mocap_constraints}
($C_i$) each point $\pointSet_{t,i}$ can be assigned to at most one label and vice versa;\label{label_disjoint}
($C_{ii}$) each point $\pointSet_{t,i}$ can be assigned to at most one tracklet;\label{traj_disjoint}
($C_{iii}$) the label \nan is an exception that can be matched to more than one point and can be present in multiple tracklets in each frame.\label{nan_unconstrained}

\section{\soma}\label{sec:method_soma}
\begin{figure*}[hbt!]
	\centering
	\includegraphics[width=\linewidth]{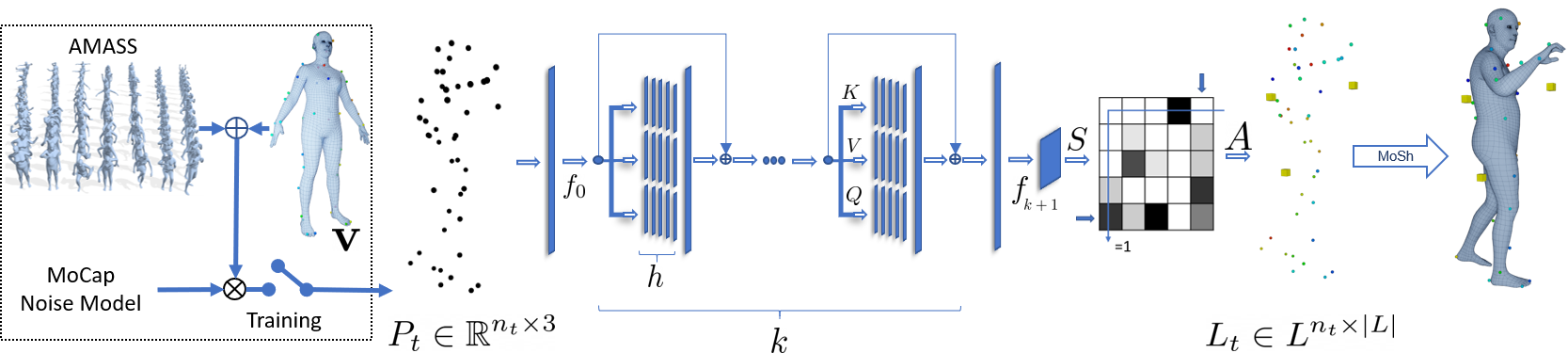}
			\vspace*{-1.5em}
	\caption{%
		We train \soma solely with synthetic data, Sec. \ref{sec:method_synthetic_data}. 
		At runtime, \soma receives an unprocessed 3D sparse
                \mocap point cloud, $P_t$, with a varying number of points. 
		These are median centered and passed through the
                pipeline, consisting of self-attention layers,
                Sec. \ref{sec:method_self_attention}, and a final
                normalization  to encourage bijective label-point correspondence, Sec. \ref{sec:method_constrained_point_labeling}. 
		The network outputs labels, $\labelSet_t$, assigned to each point, that correspond to markers in the training marker layout, $\labelSetVertexVector$, with an additional \nan label. 
		Finally, a 3D body is fit to the labeled points using \mosh, Sec. \ref{sec:method_mosh}. 
		The dimensionality of the features are 
		$\{K,V,Q\} \in \mathbb{R}^{n_t\times \frac{d_{model}}{\numAttHeads}}$, 
		$f_0 \in \mathbb{R}^{n_t\times d_{model}}$, 
		$f_{\numAttLayers+1} \in \mathbb{R}^{n_t\times 256}$,  
		$\asmat \in \mathbb{R}^{n_t\times |\labelSet|}$.
	}
	\label{fig:sc_architecture}
		\vspace*{-1.0em}
\end{figure*}

\subsection{Self-Attention on \Mocap Point Clouds}\label{sec:method_self_attention}
The \soma system pipline is summarized in Fig.~\ref{fig:sc_architecture}. 
The input to \soma is a single frame of sparse, unordered, points, the cardinality of which varies with each timestamp due to occlusions and ghost points.  
To process such data, we exploit multiple layers of self-attention \cite{vaswaniAttentionAllYou2017}, with a multi-head formulation, concatenated via residual operations \cite{resnet_2016_he, vaswaniAttentionAllYou2017}.
Multiple layers increase the capacity of the model and %
enable the network to have a local and a global view of the point cloud, which helps disambiguate points.%

We define self-attention span as the average of attention weights over random sequences picked from our validation dataset.
Figure \ref{fig:layer_label_attention_body} visualizes the attention placed on markers at the first and last self-attention layers; 
the intensity of the red color correlates with the amount of attention. 
Note that the points are shown on a body in a canonical pose but the actual \mocap point cloud data is in many poses.
Deeper layers focus attention on geodesically near body regions (wrist: upper and lower arm) or regions that are highly correlated (left foot: right foot and left knee), indicating that the network has figured out the spatial structure of the body and correlations between parts even though the observed data points are non-linearly transformed in Euclidean space by articulation.
In Appendix~\ref{supmat:self_attention_span}, we provide further computational details and demonstrate the self-attention span as a function of network depth. 
Also, we present a model-selection experiment to choose the optimum number of layers. %

\subsection{Constrained Point Labeling}\label{sec:method_constrained_point_labeling}

In the final stage of the architecture, \soma predicts a non-square score matrix $\ascorePred \in \mathbb{R}^{n_t \times \numMarkers}$. 
To satisfy the constraints $C_{i}$ and $C_{ii}$, we employ a log-domain, stable implementation of optimal transport  \cite{SuperGlue_Sarlin_2020_CVPR} described by \cite{peyre2019computational}.
The optimal transport layer depends on iterative Sinkhorn normalization \cite{adams2011ranking, sinkhorn1967, Sinkhorn1964}, which constrains rows and columns to sum to 1 for \emph{available} points and labels. 
To deal with missing markers and ghost points, following \cite{SuperGlue_Sarlin_2020_CVPR}, we introduce \emph{dustbins} by
appending an extra last row and column to the score matrix.
These can be assigned to multiple \emph{unmatched} points and labels, hence respectively summing to $n_t$ and $\numMarkers$. 
After normalization, we reach the augmented assignment matrix, $\augasmatPred \in [0,1]^{(n_t+1) \times |\labelSet|} $, from which we drop the appended row, for unmatched labels, yielding the final normalized assignment matrix  $\asmatPred \in [0,1]^{n_t \times |\labelSet|}$.

While Ghorbani et al.~\cite{ghorbanietemadetal2019} use a similar score normalization approach, their method cannot handle unmatched cases, %
which is critical to handle real \mocap point cloud data, in its raw form. %

\subsection{Solving for the Body}\label{sec:method_mosh}
Once mocap points are labeled, we ``solve'' for the articulated body that lies behind the motion.
Typical  \mocap solvers \cite{Holden:2018:Robust, 2019optitrack, 2019vicon}  fit a skeletal model to  the labeled markers.
Instead, here we fit an entire articulated 3D body mesh to markers using \moshpp  \cite{Loper:SIGASIA:2014, AMASS:2019}. 
This technique gives an animated body with a skeletal structure so nothing is lost over traditional methods while yielding a full 3D body model, consistent with other recent \mocap datasets \cite{AMASS:2019, GRAB:2020}. 
Here we fit the \smplx body model. which provides forward compatibility for datasets with hands and face captures.
For more details on \mosh we refer the reader to the original paper \citemosh.

\subsection{Synthetic \Mocap Generation}\label{sec:method_synthetic_data}
\textbf{Human Body Model.} To synthetically produce realistic \mocap training data with \gt labels, 
we leverage a gender-neutral, state of the art statistical body model, \smplx \cite{SMPL-X:2019}, that uses vertex-based linear blend skinning with learned corrective blend shapes to output the global position of $|\bodyVertex|=10,475$ vertices: 
\begin{equation}
\text{\smplx}\left( \theta_b, \beta, \gamma \right):\mathbb{R}^{| \theta_b | \times | \beta | \times | \gamma |} \rightarrow \mathbb{R}^{3N}.
\end{equation}

Here $\theta_b \in \mathbb{R}^{3(\joints+1)}$ is the axis-angle representation of the body pose where $\joints = 21$ is the number of body joints of an underlying skeleton in addition to a root joint for global rotation.
We use $\beta \in \mathbb{R}^{10}$ and $\gamma \in \mathbb{R}^{3}$ to respectively parameterize body shape and the global translation.
Compared to the original \smplx notation, here we discard  parameters that control facial expressions, face and hand poses; i.e.\  respectively $\psi, \theta_f, \theta_h$. %
We build on \smplx to enable extension of \soma to datasets with face and hand markers but SMPL-X can be converted to SMPL if needed.
For more details we  %
refer the reader to \cite{SMPL-X:2019}. 

\textbf{\Mocap Noise Model.} 
Various noise sources can influence \mocap data, namely: subject body shape, motion, marker layout and the exact placement of the markers on body, occlusions, ghost points, \mocap hardware intrinsics, and more. 
To learn a robust model, we exploit \amass \cite{AMASS:2019} %
that we refit %
with a neutral gender \smplx body model and sub-sample to a unified \unifiedFramRate.
To be robust to subject body shape we generate \amass motions for
\numSmplxBetasNeutral bodies from the CAESAR dataset \cite{CAESAR}.
Specifically, for training we take parameters from the following \mocap sub-datasets of \amass: 
CMU \cite{cmuWEB}, Transitions \cite{AMASS:2019} and Pose Prior \cite{PosePrior_Akhter:CVPR:2015}.
For validation we use HumanEva \cite{HEva_Sigal:IJCV:10b}, ACCAD \cite{ACCAD}, and TotalCapture \cite{TotalCapture_Trumble:BMVC:2017}.%

Given a marker layout of a target dataset, $\labelSetVertexVector$, as a vector of length $M$ in which the index corresponds to the maker label and the entry to a vertex on the \smplx body mesh, 
together with a vector of marker-body distances $\mathbf{d}$ %
we can place virtual markers, $\markers \in \mathbb{R}^{M\times 3}$ on the body:
\begin{eqnarray}
	\markers = \text{\smplx}\left( \theta_b, \beta, \gamma \right)|_{\labelSetVertexVector} + \mathbf{N}|_{\labelSetVertexVector} \odot \mathbf{d}.
	\label{eqn:synthetic_marker}
\end{eqnarray}
Here $\mathbf{N} \in \mathbb{R}^{V\times 3}$ is a matrix of vertex normals and %
$|_{\labelSetVertexVector}$ picks the vector of elements (vertices or normals) corresponding to vertices defined by the marker layout.

With this, we produce a library of \mocap frames and corrupt them with various controllable noise sources. 
Specifically, to generate a noisy layout, we randomly sample a vertex in the 1-ring neighborhood of the original vertex specified by the marker layout, effectively producing a different marker placement, $\labelSetVariedVertexVector$, for each data point. 
Instead of normalizing the global body orientation, common to previous methods \cite{ghorbanietemadetal2019, hanliuetal2018, Holden:2018:Robust, mocapsolver2021}, we add a random rotation $r \in [0, 2\pi]$ to the global root orientation of every body frame to augment this value.
Further, we copy the noise for each label from the real \amass \mocap markers to help generalize to \mocap hardware differences. %
We create a database of the differences between the simulated and actual markers of \amass and 
draw random samples from this noise model to add  to the synthetic marker positions.

Furthermore, we append ghost points to the generated \mocap frame, %
by drawing random samples from a 3D Gaussian distribution with mean and standard deviation equal to the median and standard deviation of the marker positions, respectively. %
Moreover, to simulate marker occlusions we take random samples from a uniform distribution representing the index of the markers and occlude selected markers by replacing their value with zero. 
The number of added ghost points and occlusions in each frame can also be subject to randomness.
\begin{figure}[tbh]
	\centering
	\includegraphics[width=\columnwidth]{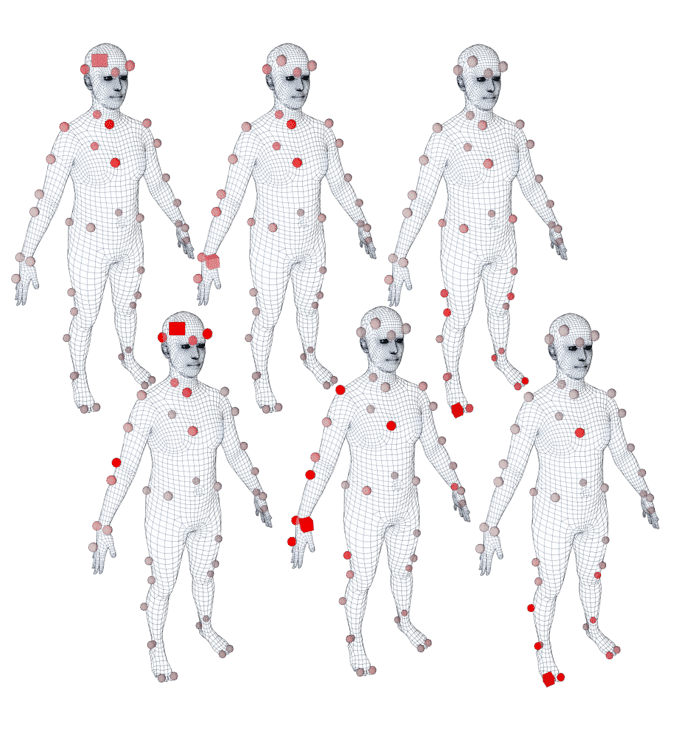}
\vspace{-.50in}
	\caption{Attention span for different markers on the body in a canonical pose. 
          The cube shows the marker of
          interest and color intensity depicts the average value of attention
          across frames of 50 randomly selected sequences. Each
          column shows a different marker.
          At the first layer (top) we see wider attention compared to the deepest layer (bottom).  
}
	\label{fig:layer_label_attention_body}
		\vspace*{-1.5em}
\end{figure}

At test time, to mimic broken trajectories of passive \mocap systems, we randomly choose a trajectory and break it at random timestamps.
To break a trajectory we copy marker values at the onset of the breakage and create a new trajectory whose previous values up-to -the breakage are zero and the rest are replaced by the marker of interest. 
The original marker trajectory after breakage is replaced by zeros.

Finally, at train and test times we randomly permute the markers to create an unordered set of 3D \mocap points. %
In contrast to \cite{ghorbanietemadetal2019}, the permutations are random and not limited to a specific set of permutations. %

\subsection{Implementation Details}\label{sec:method_implementation}

\textbf{Loss.} The total loss for training \soma is formulated as, $\loss = c_{l}\loss_{\asmat} + c_{reg}\loss_{reg}$, where: %
\begin{eqnarray}\label{eq:loss_total}
	\loss_{\asmat} & = & \frac{-1}{\sum_{i,j} \augasmatGt_{i,j}} \sum_{i,j} {\lossAsmatWeight_{i,j} \cdot \augasmatGt_{i,j} \cdot log(\augasmat_{i,j})},\\
	\loss_{reg} & = & ||\phi||_2^{2}. %
\end{eqnarray}
$\augasmat$ is the augmented assignment matrix, and $\augasmatGt$ is its ground-truth version. 
$\lossAsmatWeight$ is a matrix to down-weight the influence of the over-represented class, i.e.\ the \nan label, by the reciprocal of its occurrence frequency. 
$\loss_{reg}$ is $L_2$ regularization on the model parameters. 
In Appendix~\ref{supmat:implementation_details} we present further architecture details.

\textbf{Using \soma.}\label{sec:method_runtime_application}
The automatic labeling pipeline starts with a labeled \mocap frame that can roughly resemble the marker layout of the target dataset. 
If the dataset has significant variations in marker layout or many displaced or removed markers, 
one labeled frame per each major variation is needed.
We then train one model for the whole dataset. %
After training with synthetic data produced for the target marker layout, we apply \soma on \mocap sequences in a {\em per-frame} mode; \ie~each frame is processed independently.
On a GPU, auto-labeling runs at $52\pm 12 Hz$ in non-batched mode and, for a batch of 30 frames runtime is $1500\pm 400 Hz$. %
In cases where the \mocap hardware provides tracklets of points, we assign the most frequent label for a tracklet to all of the member points; we call this \emph{tracklet labeling.}
For detailed examples of using \soma, including a general model to facilitate labeling the initial frame, i.e.  ``label priming'', see Appendix~\ref{supmat:processing_real_mocap}.

\section{Experiments}\label{sec:experiments}
\textbf{Evaluation Datasets.}
We evaluate \soma quantitatively on various \mocap datasets with real marker data and synthetic noise; namely: BMLrub \cite{BMLrub}, BMLmovi \cite{ghorbani2020movi}, and KIT \cite{KIT_Dataset}.
The individual datasets offer various maker layouts with different marker density, subject shape variation, body pose, and recording systems. 
We take original marker data from their respective public access points and further corrupt the data with controlled noise, namely marker occlusion, ghost points, and per-frame random shuffling (Sec. \ref{sec:method_synthetic_data}).
For per-frame experiments, broken trajectory noise is not used.
We also collect a new ``\soma dataset'', which we use for direct comparison with Vicon. 

To avoid overfitting hyper-parameters to test datasets, we utilize a separate dataset for model selection and validation experiments; namely \validationDataset \citeValidationDataset, containing 215 sequences, across 4 subjects, on average using 40 markers.

\textbf{Evaluation Metrics.}
Primarily, we report mean and standard deviation of per-frame accuracy and F1 score in percentages. 
Accuracy is the proportion of correctly predicted labels over all labels, and F1 score is regarded as the harmonic-average of the precision and recall: 
\begin{equation}
	\label{eqn:f1_score}
	\text{F1} = 2 \times \frac{\text{precision} \times \text{recall}}{\text{precision} + \text{recall}},
\end{equation}
where recall is the  proportion of correct predicted labels over actual labels and precision is regarded as the proportion of actual correct labels over predicted labels.
The final F1 score for a \mocap sequence is the average of the per-frame F1 scores.

\subsection{Effectiveness of the \Mocap Noise Generation}\label{sec:experiment_noise_model_eval}
\begin{table}[]
	\centering
	\resizebox{\columnwidth}{!}{%
		\begin{tabular}{ccccccccc}
			\hline
            & \multicolumn{2}{c}{\textbf{B}}     & \multicolumn{2}{c}{\textbf{B+C}}      & \multicolumn{2}{c}{\textbf{B+G}}      & \multicolumn{2}{c}{\textbf{B+C+G}}    \\
			\multirow{-2}{*}{\textbf{\diagbox[trim=lr]{Train}{Test}}} & \cellcolor[HTML]{EFEFEF}Acc.  & F1    & \cellcolor[HTML]{EFEFEF}Acc.  & F1    & \cellcolor[HTML]{EFEFEF}Acc.  & F1    & \cellcolor[HTML]{EFEFEF}Acc.  & F1    \\ \hline
			\textbf{B}                                             & \cellcolor[HTML]{EFEFEF}97.93 & 97.37 & \cellcolor[HTML]{EFEFEF}83.55 & 81.11 & \cellcolor[HTML]{EFEFEF}86.54 & 85.97 & \cellcolor[HTML]{EFEFEF}73.95 & 71.93 \\
			\textbf{B+C}                                              & \cellcolor[HTML]{EFEFEF}97.25 & 96.31 & \cellcolor[HTML]{EFEFEF}97.21 & 96.05 & \cellcolor[HTML]{EFEFEF}95.37 & 95.29 & \cellcolor[HTML]{EFEFEF}93.79 & 92.77 \\
			\textbf{B+G}                                              & \cellcolor[HTML]{EFEFEF}98.06 & 97.19 & \cellcolor[HTML]{EFEFEF}96.14 & 94.03 & \cellcolor[HTML]{EFEFEF}97.87 & 97.65 & \cellcolor[HTML]{EFEFEF}95.27 & 93.62 \\
			\textbf{B+C+G}                                            & \cellcolor[HTML]{EFEFEF}95.74 & 94.44 & \cellcolor[HTML]{EFEFEF}95.56 & 93.91 & \cellcolor[HTML]{EFEFEF}95.73 & 95.22 & \cellcolor[HTML]{EFEFEF}95.50 & 94.44 \\ \hline
		\end{tabular}%
	}
\vspace*{-.5em}
	\caption{Per-frame labeling with synthetic noise model evaluated on real \mocap markers of \validationDataset with added noise. ``B, C, G" respectively stand for Base, Occlusion, and Ghost points. We report average accuracy and F1 scores. %
	}
	\label{tab:noise_model_eval}
\end{table}
Here we train and test \soma with various amounts of synthetic noise.
The training data is synthetically produced for the layout of \validationDataset as described in Sec.~\ref{sec:method_synthetic_data}.
We test on original markers of \valds corrupted with synthetic noise.
\emph{B} stands for no noise, \emph{B+C} for up-to 5 markers occluded per-frame, \emph{B+G} for up-to 3 ghost points per-frame, and \emph{B+C+G} for the full treatment of the occlusion and ghost point noise. 
Table~\ref{tab:noise_model_eval} shows different  training and testing scenarios.
In general, matching the model training noise to the true noise produces the best results but training with the full noise model (B+C+G) gives competitive performance and is a good choice when the noise level is unknown.
Training with more noise improves robustness.

\subsection{Comparison with Previous Work}\label{sec:expriment_comparision_previous_work}
\begin{table}[]
	\centering
	\resizebox{\columnwidth}{!}{%
		\begin{tabular}{lccccccc}
			\hline
			\multicolumn{1}{c}{\multirow{2}{*}{Method}}         & \multicolumn{7}{c}{Number of Exact Per-Frame Occlusions} \\ \cline{2-8} 
			\multicolumn{1}{c}{}                                & 0      & 1      & 2      & 3     & 4     & 5     & 5+G   \\ \hline
			Holzreiter et al. \cite{holzreiter2005autolabeling} & 88.16  & 79.00  & 72.42  & 67.16 & 61.13 & 52.10 & ---   \\
			Maycock et al. \cite{maycockrohlingetal2015}        & 83.19  & 79.35  & 76.44  & 74.91 & 71.17 & 65.83 & ---   \\
			Ghorbani et al. \cite{ghorbanietemadetal2019}       & 97.11  & 96.56  & 96.13  & 95.87 & 95.75 & 94.90 & ---   \\
			\textbf{\soma-Real} & \textbf{99.08} & \textbf{98.97} & \textbf{98.85} & \textbf{98.68} & \textbf{98.48} & \textbf{98.22} & \textbf{98.29} \\
			\textbf{\soma-Synthetic}                            & 99.16  & 98.92  & 98.54  & 98.17 & 97.61 & 97.07 & 95.13 \\
			\textbf{\soma\/*}                                    & 98.38  & 98.28  & 98.17  & 98.03 & 97.86 & 97.66 & 97.56 \\ \hline
		\end{tabular}%
	}
			\vspace*{-.5em}
	\caption{Comparing \soma with prior work on the same data. We train \soma in three different ways: once with \emph{real} data; once with \emph{synthetic} markers placed on bodies of the same motions obtained from \amass; and ultimately once with the data produced in Sec. \ref{sec:method_synthetic_data}, designated with \emph{*}. See Fig.~\ref{tab:comparision_previous_work_std} for standard deviations.}
	\label{tab:comparision_previous_work}
				\vspace*{-1.5em}
\end{table}

We compare \soma to %
prior work in Tab.~\ref{tab:comparision_previous_work} under the same conditions.
Specifically, we use train and test data splits of the BMLrub dataset explained by \cite{ghorbanietemadetal2019}.  
The test happens on real markers with synthetic noise. We train \soma once with real marker data and once by synthetic markers produced by motions of the same split.
Additionally, we train \soma with the full synthetic data outlined in Sec.~\ref{sec:method_synthetic_data}. %
The performance of other competing methods is as reported by \cite{ghorbanietemadetal2019}.  
All versions of \soma outperform previous methods.
The model (\soma\/*) trained with synthetic markers and varied training poses from \amass is competitive with
the models trained on limited real data or synthetic marker data with dataset-specific motions.
This is likely due to the rich variation in our noise generation pipeline.
In contrast to prior work, \soma is robust to increasing occlusion and can process ghost points without extra heuristics; i.e.\ last column in Tab.~\ref{tab:comparision_previous_work}.

\subsection{Performance Robustness}\label{sec:experiments_mocap_setup_variation}
\begin{table*}[]
	\centering
	\resizebox{\textwidth}{!}{%
		\begin{tabular}{ccccccclc}
			\hline
			& \multicolumn{2}{c}{Per-Frame}                               & \multicolumn{2}{c}{Tracklet}                                &                                                                           &                                                                           & \multicolumn{1}{c}{}                                                                        &                                                                            \\
			\multirow{-2}{*}{Datasets}     & \cellcolor[HTML]{EFEFEF}Acc.             & F1               & \cellcolor[HTML]{EFEFEF}Acc.             & F1               & \multirow{-2}{*}{\begin{tabular}[c]{@{}c@{}}$\#$ \\ Markers\end{tabular}} & \multirow{-2}{*}{\begin{tabular}[c]{@{}c@{}}$\#$ \\ Motions\end{tabular}} & \multicolumn{1}{c}{\multirow{-2}{*}{\begin{tabular}[c]{@{}c@{}}$\#$\\ Frames\end{tabular}}} & \multirow{-2}{*}{\begin{tabular}[c]{@{}c@{}}$\#$ \\ Subjects\end{tabular}} \\ \hline
			BMLrub \cite{BMLrub}           & \cellcolor[HTML]{EFEFEF}98.15 $\pm$ 2.78 & 97.75 $\pm$ 3.23 & \cellcolor[HTML]{EFEFEF}98.77 $\pm$ 1.58 & 98.65 $\pm$ 1.89 & 41                                                                        & 3013                                                                      & 3757725                                                                                     & 111                                                                        \\
			KIT\cite{KIT_Dataset}          & \cellcolor[HTML]{EFEFEF}94.97 $\pm$ 2.42 & 95.51 $\pm$ 2.65 & \cellcolor[HTML]{EFEFEF}95.46 $\pm$ 1.87 & 97.10 $\pm$ 2.00 & 53                                                                        & 3884                                                                      & 3504524                                                                                     & 48                                                                         \\ 
			BMLmovi\cite{ghorbani2020movi} & \cellcolor[HTML]{EFEFEF}95.90 $\pm$ 4.65 & 95.12 $\pm$ 5.26 & \cellcolor[HTML]{EFEFEF}97.33 $\pm$ 2.29 & 96.87 $\pm$ 2.60 & 67                                                                        & 1863                                                                      & 1255447                                                                                     & 89                                                                         \\ \hline
		\end{tabular}%
	}
	\vspace*{-.5em}
	\caption{Performance of \soma on real marker data of various datasets with large variation in number of subjects, body pose, markers, and hardware specifics. We corrupt the real marker data with additional noise, and forget the labels, turning it into a raw \mpc before passing through \soma pipeline.}
	\label{tab:performance_various_datasets}
	\vspace*{-.5em}
\end{table*}

\textbf{Performance on Various \Mocap Datasets} could vary due to variations in the marker density, \mocap quality, subject shape and motions. 
To assess such cases we take three full scale \mocap datasets and corrupt the markers with synthetic noise including up to 50 broken trajectories that best mimic the situation with a realistic unlabeled \mocap scenario. 
Additionally we evaluate tracklet labeling explained in Sec.~\ref{sec:method_runtime_application}. %
Table \ref{tab:performance_various_datasets} shows consistent high performance of \soma across the datasets. 
When tracklets are available, as they often are, tracklet labeling improves performance across the board.

\textbf{Performance on Subsets or Supersets of a Specific Marker Layout} could vary since this introduces ``structured'' noise.
A superset marker layout is the set of all labels in a dataset, which may combine multiple marker layouts.
A \emph{base model} trained on a superset marker layout and tested on subsets would be subject to structured occlusions, while a model trained on subset and tested on the superset \emph{base \mocap} would see structured ghost points. 
These situations commonly happen across a dataset when trial coordinators improvise on the planned marker layout for a special take with additional or fewer markers. 
Alternatively, markers often fall off  during the trial. 
To quantitatively determine the range of performance variance we take the marker layout of the validation dataset, \validationDataset, and omit markers in progressive steps; Tab.~\ref{tab:performance_mlayout_variation}.
The model, trained on subset layout and tested on base markers (superset), shows greater deterioration in performance than the base model trained on the superset and tested on reduced marker sets.
\begin{table*}[]
	\centering
	\resizebox{\textwidth}{!}{%
		\begin{tabular}{ccccccccc}
			\hline
            & \multicolumn{2}{c}{3}                                       & \multicolumn{2}{c}{5}                                       & \multicolumn{2}{c}{6}                                       & \multicolumn{2}{c}{12}                                       \\

			\multirow{-2}{*}{\begin{tabular}[c]{@{}c@{}}$\#$ Markers \\ Removed\end{tabular}} & \cellcolor[HTML]{EFEFEF}Acc.             & F1               & \cellcolor[HTML]{EFEFEF}Acc.             & F1               & \cellcolor[HTML]{EFEFEF}Acc.             & F1               & \cellcolor[HTML]{EFEFEF}Acc.              & F1               \\ \hline
			Base Model                                                                        & \cellcolor[HTML]{EFEFEF}95.35 $\pm$ 6.52 & 94.40 $\pm$ 7.37 & \cellcolor[HTML]{EFEFEF}94.08 $\pm$ 7.04 & 91.75 $\pm$ 8.32 & \cellcolor[HTML]{EFEFEF}93.41 $\pm$ 7.84 & 90.73 $\pm$ 9.33 & \cellcolor[HTML]{EFEFEF}88.25 $\pm$ 14.04 & 8.80 $\pm$ 16.07 \\
		Base MoCap                                                                        & \cellcolor[HTML]{EFEFEF}91.78 $\pm$ 10.13 & 90.68 $\pm$ 10.90 & \cellcolor[HTML]{EFEFEF}90.73 $\pm$ 9.38 & 90.12 $\pm$ 9.96 & \cellcolor[HTML]{EFEFEF}91.89 $\pm$ 8.97 & 91.00 $\pm$ 9.71 & \cellcolor[HTML]{EFEFEF}87.46 $\pm$ 10.77 & 86.78 $\pm$ 12.00 \\ \hline
		\end{tabular}%
	}
\vspace{-0.5em}
	\caption{Robustness to variations in marker layout. 
		First row: A \emph{base model} is trained with full marker layout
                (superset) and tested per-frame on real markers from
                the validation set (\validationDataset) with omitted
                markers (subset). 
		Second row: One model is trained per varied layout
                (subset) and tested on \emph{base \mocap} markers (superset). }
	\label{tab:performance_mlayout_variation}
\vspace*{-1.em}
\end{table*}

\subsection{Ablation Studies}\label{sec:experiments_ablation}
\begin{table}[]
	\centering
	\resizebox{\columnwidth}{!}{%
		\begin{tabular}{l
				>{\columncolor[HTML]{EFEFEF}}c c}
			\hline
			\multicolumn{1}{c}{\cellcolor[HTML]{EFEFEF}{\color[HTML]{333333} Version}} & Accuracy                                  & F1                \\ \hline
			Base                                                                       & 95.50 $\pm$ 5.33                          & 94.66 $\pm$ 6.03  \\
			\quad - AMASS Noise Model                                                    & 94.73$\pm$ 5.52                           & 93.73 $\pm$ 6.25  \\
			\quad - CAESAR bodies                                                        & \cellcolor[HTML]{EFEFEF}95.21 $\pm$ 6.83  & 94.31 $\pm$ 7.57  \\
			\quad - Log-Softmax Instead of Sinkhorn                                      & \cellcolor[HTML]{EFEFEF}91.51 $\pm$ 10.69 & 90.10 $\pm$ 11.47 \\
			\quad - Random Marker Placement                                              & \cellcolor[HTML]{EFEFEF}89.41 $\pm$ 8.06  & 87.78 $\pm$ 8.85  \\
			\quad - Transformer                                                          & 11.36 $\pm$ 6.54                          & 7.54 $\pm$ 6.22   \\ \hline
		\end{tabular}%
	}
\vspace*{-.5em}
	\caption{Ablation study of \soma components on the \valds
          dataset. The numbers reﬂect the contribution of each
          component in overall per-frame performance of \soma. 
          We take the full base model and remove one component at a time.
}
	\label{tab:ablation_study}
	\vspace*{-.5em}
\end{table}

Table \ref{tab:ablation_study} shows effect of various components %
in the final performance of \soma on the validation dataset, \validationDataset.
The self-attention layers %
and the novel random marker placement noise  %
play the most significant role in overall performance of the model. 
The optimal transport layer marginally improves accuracy of the model compared to %
the Log-Softmax normalization.

\subsection{Application}\label{sec:experiments_application}

\begin{table}[]
	\centering
	\resizebox{\columnwidth}{!}{%
		\begin{tabular}{ccccc}
			\hline
			\textbf{} & $V2V_{mm}^{mean}$ & $V2V_{mm}^{median}$ & Acc.             & F1               \\ \hline
			\shogun   & 0.00 $\pm$ 0.11   & 0.00                & 100.0 $\pm$ 0.00 & 100.0 $\pm$ 0.00 \\
			\soma     & 0.08 $\pm$ 2.09   & 0.00                & 99.94 $\pm$ 0.47 & 99.92 $\pm$ 0.64 \\ \hline
		\end{tabular}%
	}
\vspace*{-.8em}
	\caption{\soma vs \shogun. 
		On a manually labeled dataset with passive markers, we compare \soma against a commercial tool for labeling performance and surface reconstruction.}
	\label{tab:against_vicon}
	\vspace*{-0.5em}
\end{table}
\begin{table}[]
	\centering
	\resizebox{\columnwidth}{!}{%
		\begin{tabular}{@{}cccccc@{}}
			\toprule
			\textbf{} & Type & $\#$ Points & $\#$ Subjects & Minutes & Success Ratio \\ \midrule
			CMU-II \cite{cmuWEB}                  & P & 40-255 & 41 & 116.30 & 80.0$\%$   \\
			DanceDB \cite{DanceDB:2018:aristidou} & A & 38     & 20 & 203.38 & 81.26$\%$  \\
			Mixamo \cite{mixamoWEB}               & A & 38-96  & 29 & 195.37 & 78.31$\%$  \\
			\soma                                 & P & 53-140 & 2  & 18.27  & 100.00$\%$ \\
			\rowcolor[HTML]{EFEFEF} 
			Total                                 &   &        &    & 533.32 &            \\ \bottomrule
		\end{tabular}%
	}
		\vspace{-.5em}
		\caption{Processing uncleaned, unlabeled \mocap datasets with \soma. 
		Input to the pipeline are \mocap sequences with
                possibly varying number of points; \soma labels the
                points as markers and afterwards \mosh is applied on
                the data to solve for the body surface. 
	P and A stand for passive and active marker systems respectively.}
		\label{tab:processed_datasets}
\vspace{-1.5em}
\end{table}

\textbf{Comparison with a Commercial Tool.} 
To compare directly with \vicon's \shogun auto-labeling tool, we record a new ``\soma'' \mocap dataset with $2$ subjects, performing $11$ motion types, including dance, clap, kick, etc.,  using a \vicon system with $54$ infrared ``Vantage $16$'' \cite{2019vicon} cameras operating at $120$ Hz. 
In total, we record $69$ motions and intentionally use a marker layout preferred by \vicon. 
We manually label this dataset and treat these labels as \gt.  
We process the reconstructed \mocap point cloud data with both \soma (using tracklet labeling) and with \shogun. %
The results in Tab.~\ref{tab:against_vicon} show that \soma achieves sub-millimeter accuracy and similar performance compared with the propriety tool while not requiring subject calibration data. 
In Tab.~\ref{tab:soma_dataset_details} we present further details of this dataset.

\textbf{Processing Real \Mocap Datasets} with different capture technologies and unknown subject calibration is presented in Tab. ~\ref{tab:processed_datasets}. 
For each dataset, \soma is trained on the marker superset using only synthetic data.
\soma effectively enables running \moshpp on \mocap point cloud data to extract realistic bodies. 
The results are not perfect and we  manually remove sequences that do not pass a subjective quality bar (see Appendix~\ref{supmat:processing_real_mocap} and the accompanying video for examples).
Table ~\ref{tab:processed_datasets} indicates the percentage of successful minutes of \mocap.  
Failures are typically due to poor \mocap quality. 
Note that the \soma dataset is very high quality with many cameras and, here, the success rate is 100\%.
For sample renders refer to the accompanying video.

\vspace*{-.4em}
\section{Conclusion}\label{sec:conclusion}
\soma addresses the problem of robustly labeling of raw \mocap point cloud sequences of human bodies in motion, subject to noise and variations across subjects, motions, marker placement, marker density, \mocap quality, and capture technology. 
\soma solves this problem using several innovations including a novel self-attention mechanism and a matching component that deals with outliers and missing data.  
We train \soma end-to-end on synthetic data using several techniques to add realistic noise that enable generalization to real data.
We extensively validate the performance of \soma showing that it is more accurate than previous research methods and comparable in accuracy to a commercial system while being significantly more flexible.
\soma is also freely available for research purposes.

\textbf{Limitations and Future Work.} 
\soma performs per-frame \mpc  labeling and, hence, does not exploit temporal information.
A temporal model could potentially improve accuracy.
As with any learning-based method, \soma may be limited in generalizing to new motions outside the training data.
Using AMASS, however, the variability of the training data is large and we did not observe problems with generalization.
By exploiting the full \smplx body model in synthetic data generation pipeline we plan to extend the method to label hand and face markers.
Relying on feed-forward components, \soma is extremely fast and coupled with a suitable \mocap solver could potentially recover bodies in real-time from \mocap point clouds.

\vspace*{-.1em}

	\noindent
	{\small
		{\bf   Acknowledgments}: We thank Senya Polikovsky, Markus Höschle, Galina Henz (GH), and Tobias Bauch for the \mocap facility. 
		We thank Alex Valisu, Elisha Denham, Leyre Sánchez Viñuela, Felipe Mattioni and Jakob Reinhardt for \mocap cleaning. 
		We thank GH, and Tsvetelina Alexiadis for trial coordination.
		We thank Benjamin Pellkofer and Jonathan Williams for website developments.
		{\bf   Disclosure:}
\url{https://files.is.tue.mpg.de/black/CoI/ICCV2021.txt}
	}

{\small
	\bibliographystyle{ieee_fullname}
	\bibliography{soma}

\begin{thebibliography}{10}\itemsep=-1pt

\bibitem{falcon2019pytorch}
Pytorch lightning, 2019.

\bibitem{Abdulkader_1998}
Mohammad~Abdulkader Abdulrahim.
\newblock {\em Parallel Algorithms for Labeled Graph Matching}.
\newblock PhD thesis, Colorado School of Mines, USA, 1998.
\newblock AAI0599838.

\bibitem{adams2011ranking}
Ryan~Prescott Adams and Richard~S. Zemel.
\newblock Ranking via {Sinkhorn Propagation}, 2011.

\bibitem{PosePrior_Akhter:CVPR:2015}
Ijaz Akhter and Michael~J. Black.
\newblock Pose-conditioned joint angle limits for {3D} human pose
  reconstruction.
\newblock In {\em CVPR}, pages 1446--1455, 2015.

\bibitem{DanceDB:2018:aristidou}
Andreas Aristidou, Efstathios Stavrakis, Margarita Papaefthimiou, George
  Papagiannakis, and Yiorgos Chrysanthou.
\newblock Style-based motion analysis for dance composition.
\newblock {\em The Visual Computer}, 34:1725–1737, 2018.

\bibitem{berg_2005_shape_matching}
Alexander~C. Berg, Tamara~L. Berg, and Jitendra Malik.
\newblock Shape matching and object recognition using low distortion
  correspondences.
\newblock In {\em CVPR}, volume~1, pages 26--33, 2005.

\bibitem{Caetano_2007_graph_matching}
Tibério~S. Caetano, Julian~J. McAuley, Li Cheng, Quoc~V. Le, and Alex~J.
  Smola.
\newblock Learning graph matching.
\newblock In {\em ICCV}, pages 1--8, 2007.

\bibitem{pointnet_charles_2017}
R.~Qi Charles, Hao Su, Mo Kaichun, and Leonidas~J. Guibas.
\newblock Pointnet: Deep learning on point sets for {3D} classification and
  segmentation.
\newblock In {\em CVPR}, pages 77--85, 2017.

\bibitem{mocapsolver2021}
Kang Chen, Yupan Wang, Song-Hai Zhang, Sen-Zhe Xu, Weidong Zhang, and Shi-Min
  Hu.
\newblock {MoCap-Solver}: A neural solver for optical motion capture data.
\newblock {\em ACM Transactions on Graphics (TOG)}, 40(4), 2021.

\bibitem{contini1972}
Renato Contini.
\newblock Body segment parameters, part {II}.
\newblock {\em Artificial Limbs}, 16(1):1--19, 1972.

\bibitem{mixamoWEB}
Adobe Mixamo~MoCap Dataset, 2019.

\bibitem{cmuWEB}
Carnegie Mellon University ({CMU})~MoCap Dataset, 2019.

\bibitem{ACCAD}
Advanced Computing~Center for~the Arts and Design ({ACCAD})~MoCap Dataset,
  2019.

\bibitem{ge_cvpr16_3d}
Liuhao Ge, Hui Liang, Junsong Yuan, and Daniel Thalmann.
\newblock Robust {3D} hand pose estimation in single depth images: From
  single-view {CNN} to multi-view {CNN}s.
\newblock In {\em CVPR}, pages 3593--3601, 2016.

\bibitem{ghorbanietemadetal2019}
Saeed Ghorbani, Ali Etemad, and Nikolaus~F. Troje.
\newblock Auto-labelling of markers in optical motion capture by permutation
  learning.
\newblock In {\em Advances in Computer Graphics}, pages 167--178, Cham, 2019.
  Springer International Publishing.

\bibitem{ghorbani2020movi}
Saeed Ghorbani, Kimia Mahdaviani, Anne Thaler, Konrad Kording, Douglas~James
  Cook, Gunnar Blohm, and Nikolaus~F. Troje.
\newblock {MoVi}: A large multi-purpose human motion and video dataset.
\newblock {\em PLOS ONE}, 16(6):1--15, 2021.

\bibitem{hanliuetal2018}
Shangchen Han, Beibei Liu, Robert Wang, Yuting Ye, Christopher~D. Twigg, and
  Kenrick Kin.
\newblock Online optical marker-based hand tracking with deep labels.
\newblock {\em ACM Transactions on Graphics ({TOG})}, 37(4):1--10, 2018.

\bibitem{resnet_2016_he}
Kaiming He, Xiangyu Zhang, Shaoqing Ren, and Jian Sun.
\newblock Deep residual learning for image recognition.
\newblock In {\em CVPR}, pages 770--778, 2016.

\bibitem{herdafuaetal2001}
Lorna Herda, Pascal Fua, Ralf Pl{\"a}nkers, Ronan Boulic, and Daniel Thalmann.
\newblock Using skeleton-based tracking to increase the reliability of optical
  motion capture.
\newblock {\em Human Movement Science}, 20(3):313--341, 2001.

\bibitem{Holden:2018:Robust}
Daniel Holden.
\newblock Robust solving of optical motion capture data by denoising.
\newblock {\em ACM Transactions on Graphics ({TOG})}, 37(4):165:1--165:12,
  2018.

\bibitem{holzreiter2005autolabeling}
Stefan Holzreiter.
\newblock Autolabeling 3d tracks using neural networks.
\newblock {\em Clinical Biomechanics}, 20(1):1--8, 2005.

\bibitem{2019optitrack}
NaturalPoint{,} Inc.
\newblock Optitrack motion capture systems, 2019.

\bibitem{2019phasespace}
PhaseSpace{,} Inc., 2019.

\bibitem{batchNorm2015}
Sergey Ioffe and Christian Szegedy.
\newblock Batch normalization: Accelerating deep network training by reducing
  internal covariate shift.
\newblock In {\em ICML}, page 448–456, 2015.

\bibitem{FRANK2018}
Hanbyul Joo, Tomas Simon, and Yaser Sheikh.
\newblock Total capture: A {3D} deformation model for tracking faces, hands,
  and bodies.
\newblock In {\em CVPR}, pages 8320--8329, 2018.

\bibitem{joukovetal2020}
Vladimir Joukov, Jonathan F.~S. Lin, Kevin Westermann, and Dana Kuli{\'{c}}.
\newblock Real-time unlabeled marker pose estimation via constrained extended
  kalman filter.
\newblock In {\em Proceedings of the International Symposium on Experimental
  Robotics}, pages 762--771, Cham, 2020. Springer International Publishing.

\bibitem{kingmaba2014}
Diederik Kingma and Jimmy Ba.
\newblock Adam: A method for stochastic optimization.
\newblock In {\em ICLR}, 2015.

\bibitem{Leordeanu_2005}
Marius Leordeanu and Martial Hebert.
\newblock A spectral technique for correspondence problems using pairwise
  constraints.
\newblock In {\em ICCV}, page 1482–1489, USA, 2005. IEEE Computer Society.

\bibitem{Loper:SIGASIA:2014}
Matthew Loper, Naureen Mahmood, and Michael~J. Black.
\newblock {MoSh}: Motion and shape capture from sparse markers.
\newblock {\em ACM Transactions on Graphics ({TOG})}, 33(6):1--13, 2014.

\bibitem{2019vicon}
Vicon Motion~Systems{,} Ltd.
\newblock Motion {{Capture Systems}}, 2019.

\bibitem{AMASS:2019}
Naureen Mahmood, Nima Ghorbani, Nikolaus~F. Troje, Gerard Pons-Moll, and
  Michael~J. Black.
\newblock {AMASS}: Archive of motion capture as surface shapes.
\newblock In {\em ICCV}, pages 5441--5450, 2019.

\bibitem{KIT_Dataset}
Christian Mandery, \"Omer Terlemez, Martin Do, Nikolaus Vahrenkamp, and Tamim
  Asfour.
\newblock The {KIT} whole-body human motion database.
\newblock In {\em ICAR}, pages 329--336, 2015.

\bibitem{voxnet_iros2015_maturana}
Daniel Maturana and Sebastian Scherer.
\newblock {VoxNet}: A {3D} convolutional neural network for real-time object
  recognition.
\newblock In {\em IEEE/RSJ International Conference on Intelligent Robots and
  Systems ({IROS})}, pages 922--928, 2015.

\bibitem{maycockrohlingetal2015}
Jonathan Maycock, Tobias Rohlig, Matthias Schroder, Mario Botsch, and Helge
  Ritter.
\newblock Fully automatic optical motion tracking using an inverse kinematics
  approach.
\newblock In {\em IEEE-RAS International Conference on Humanoid Robots
  (Humanoids)}, pages 461--466, 2015.

\bibitem{MPI_HDM05}
Michael Clausen Bernhard Eberhardt Bj\"{o}rn Kr\"{u}ger Andreas~Weber
  Meinard~M\"{u}ller, Tido~R\"{o}der.
\newblock Documentation mocap database {HDM05}.
\newblock Technical Report CG-2007-2, Universit\"{a}t Bonn, 2007.

\bibitem{meyerkudereretal2014}
Johannes Meyer, Markus Kuderer, Jörg Müller, and Wolfram Burgard.
\newblock Online marker labeling for fully automatic skeleton tracking in
  optical motion capture.
\newblock In {\em ICRA}, pages 5652--5657, 2014.

\bibitem{pytorch_panze_2019}
Adam Paszke, Sam Gross, Francisco Massa, Adam Lerer, James Bradbury, Gregory
  Chanan, Trevor Killeen, Zeming Lin, Natalia Gimelshein, Luca Antiga, Alban
  Desmaison, Andreas Kopf, Edward Yang, Zachary DeVito, Martin Raison, Alykhan
  Tejani, Sasank Chilamkurthy, Benoit Steiner, Lu Fang, Junjie Bai, and Soumith
  Chintala.
\newblock Pytorch: An imperative style, high-performance deep learning library.
\newblock pages 8024--8035. Curran Associates, Inc., 2019.

\bibitem{SMPL-X:2019}
Georgios Pavlakos, Vasileios Choutas, Nima Ghorbani, Timo Bolkart, Ahmed~A.
  Osman, Dimitrios Tzionas, and Michael~J. Black.
\newblock Expressive body capture: {3D} hands, face, and body from a single
  image.
\newblock In {\em CVPR}, pages 10967--10977, 2019.

\bibitem{peyre2019computational}
Gabriel Peyr{\'e} and Marco Cuturi.
\newblock Computational optimal transport.
\newblock {\em Foundations and Trends{\textregistered} in Machine Learning},
  11(5-6):355--607, 2019.

\bibitem{pointnetpp_2017}
Charles~Ruizhongtai Qi, Li Yi, Hao Su, and Leonidas~J Guibas.
\newblock Pointnet++: Deep hierarchical feature learning on point sets in a
  metric space.
\newblock In {\em NIPS}, volume~30. Curran Associates, Inc., 2017.

\bibitem{ringerlasenby2002}
Maurice Ringer and Joan Lasenby.
\newblock Multiple hypothesis tracking for automatic optical motion capture.
\newblock In {\em ECCV}, pages 524--536, Berlin, Heidelberg, 2002. Springer
  Berlin Heidelberg.

\bibitem{Ringer2002}
Maurice Ringer and Joan Lasenby.
\newblock A procedure for automatically estimating model parameters in optical
  motion capture.
\newblock {\em Image and Vision Computing}, 22(10):843--850, 2004.
\newblock British Machine Vision Conference ({BMVC}).

\bibitem{CAESAR}
Kathleen Robinette, Sherri Blackwell, Hein A.~M. Daanen, Mark Boehmer, Scott
  Fleming, Tina Brill, David Hoeferlin, and Dennis Burnsides.
\newblock Civilian american and european surface anthropometry resource
  ({CAESAR}) final report.
\newblock Technical Report AFRL-HE-WP-TR-2002-0169, US Air Force Research
  Laboratory, 2002.

\bibitem{SuperGlue_Sarlin_2020_CVPR}
Paul-Edouard Sarlin, Daniel DeTone, Tomasz Malisiewicz, and Andrew Rabinovich.
\newblock Superglue: Learning feature matching with graph neural networks.
\newblock In {\em CVPR}, pages 4937--4946, 2020.

\bibitem{schubertgkogkidisetal2015}
Tobias Schubert, Alexis Gkogkidis, Tonio Ball, and Wolfram Burgard.
\newblock Automatic initialization for skeleton tracking in optical motion
  capture.
\newblock In {\em ICRA}, pages 734--739, 2015.

\bibitem{HEva_Sigal:IJCV:10b}
Leonid Sigal, Alexandru~O. Balan, and Michael~J. Black.
\newblock {HumanEva}: Synchronized video and motion capture dataset and
  baseline algorithm for evaluation of articulated human motion.
\newblock {\em IJCV}, 87(4):4--27, 2010.

\bibitem{Sinkhorn1964}
Richard Sinkhorn.
\newblock A relationship between arbitrary positive matrices and doubly
  stochastic matrices.
\newblock {\em The Annals of Mathematical Statistics}, 35(2):876--879, 1964.

\bibitem{sinkhorn1967}
Richard Sinkhorn and Paul Knopp.
\newblock Concerning nonnegative matrices and doubly stochastic matrices.
\newblock {\em Pacific Journal of Mathematics}, 21(2):343--348, 1967.

\bibitem{Song2003}
Yang Song, Luis Goncalves, and Pietro Perona.
\newblock Unsupervised learning of human motion.
\newblock {\em TPAMI}, 25(7):814–827, 2003.

\bibitem{Steinbring2016}
Jannik Steinbring, Christian Mandery, Florian Pfaff, Florian Faion, Tamim
  Asfour, and Uwe~D. Hanebeck.
\newblock Real-time whole-body human motion tracking based on unlabeled
  markers.
\newblock In {\em IEEE International Conference on Multisensor Fusion and
  Integration for Intelligent Systems ({MFI})}, pages 583--590, 2016.

\bibitem{su15mvcnn}
Hang Su, Subhransu Maji, Evangelos Kalogerakis, and Erik~G. Learned{-}Miller.
\newblock Multi-view convolutional neural networks for {3D} shape recognition.
\newblock In {\em ICCV}, 2015.

\bibitem{GRAB:2020}
Omid Taheri, Nima Ghorbani, Michael~J. Black, and Dimitrios Tzionas.
\newblock Grab: A dataset of whole-body human grasping of objects.
\newblock In {\em ECCV}, pages 581--600, Cham, 2020. Springer International
  Publishing.

\bibitem{Torresani_2008}
Lorenzo Torresani, Vladimir Kolmogorov, and Carsten Rother.
\newblock Feature correspondence via graph matching: Models and global
  optimization.
\newblock In {\em ECCV}, pages 596--609, Berlin, Heidelberg, 2008. Springer
  Berlin Heidelberg.

\bibitem{BMLrub}
Nikolaus~F. Troje.
\newblock Decomposing biological motion: A framework for analysis and synthesis
  of human gait patterns.
\newblock {\em Journal of Vision}, 2(5):2--2, 2002.

\bibitem{TotalCapture_Trumble:BMVC:2017}
Matthew Trumble, Andrew Gilbert, Charles Malleson, Adrian Hilton, and John
  Collomosse.
\newblock Total capture: {3D} human pose estimation fusing video and inertial
  sensors.
\newblock In {\em BMVC}, pages 14.1--14.13, 2017.

\bibitem{vaswaniAttentionAllYou2017}
Ashish Vaswani, Noam Shazeer, Niki Parmar, Jakob Uszkoreit, Llion Jones,
  Aidan~N. Gomez, undefinedukasz Kaiser, and Illia Polosukhin.
\newblock Attention is all you need.
\newblock In {\em NIPs}, page 6000–6010, Red Hook, NY, USA, 2017. Curran
  Associates Inc.

\bibitem{cedric_2008_optimal_transport}
Cédric Villani.
\newblock {\em Optimal transport -- Old and New}, volume 338.
\newblock Springer Science and Business Media, 2008.

\bibitem{NonLocal2018}
Xiaolong Wang, Ross Girshick, Abhinav Gupta, and Kaiming He.
\newblock Non-local neural networks.
\newblock In {\em CVPR}, pages 7794--7803, 2018.

\bibitem{zhirongwusongetal2015b}
Zhirong Wu, Shuran Song, Aditya Khosla, Fisher Yu, Linguang Zhang, Xiaoou Tang,
  and Jianxiong Xiao.
\newblock {3D ShapeNets}: {A} deep representation for volumetric shapes.
\newblock In {\em CVPR}, pages 1912--1920, 2015.

\bibitem{GHUM2020}
Hongyi Xu, Eduard~Gabriel Bazavan, Andrei Zanfir, William~T. Freeman, Rahul
  Sukthankar, and Cristian Sminchisescu.
\newblock {GHUM \& GHUML}: Generative 3d human shape and articulated pose
  models.
\newblock In {\em CVPR}, pages 6183--6192, 2020.

\end{thebibliography}
}

\clearpage
\appendix

{\noindent\Large\textbf{Appendix}}
\counterwithin{figure}{section}
\counterwithin{table}{section}

\soma labels raw and noisy ``\mocap point clouds'' at scale, without requiring subject calibration, and across various capture technologies. 
Here we provide more details as referenced in the main paper.
Additionally, we encourage the reader to watch the {\bf supplementary video}.
Since \mocap is inherently about motion, it is very difficult to convey the quality of the results in a static format.
The video provides a much clearer picture of what \soma does and the quality of the results.
All supplementary material can be accessed in the project website \url{https://soma.is.tue.mpg.de/}

\section{Self-Attention Span}\label{supmat:self_attention_span}
\begin{figure}[h]
	\centering
	\includegraphics[width=\linewidth]{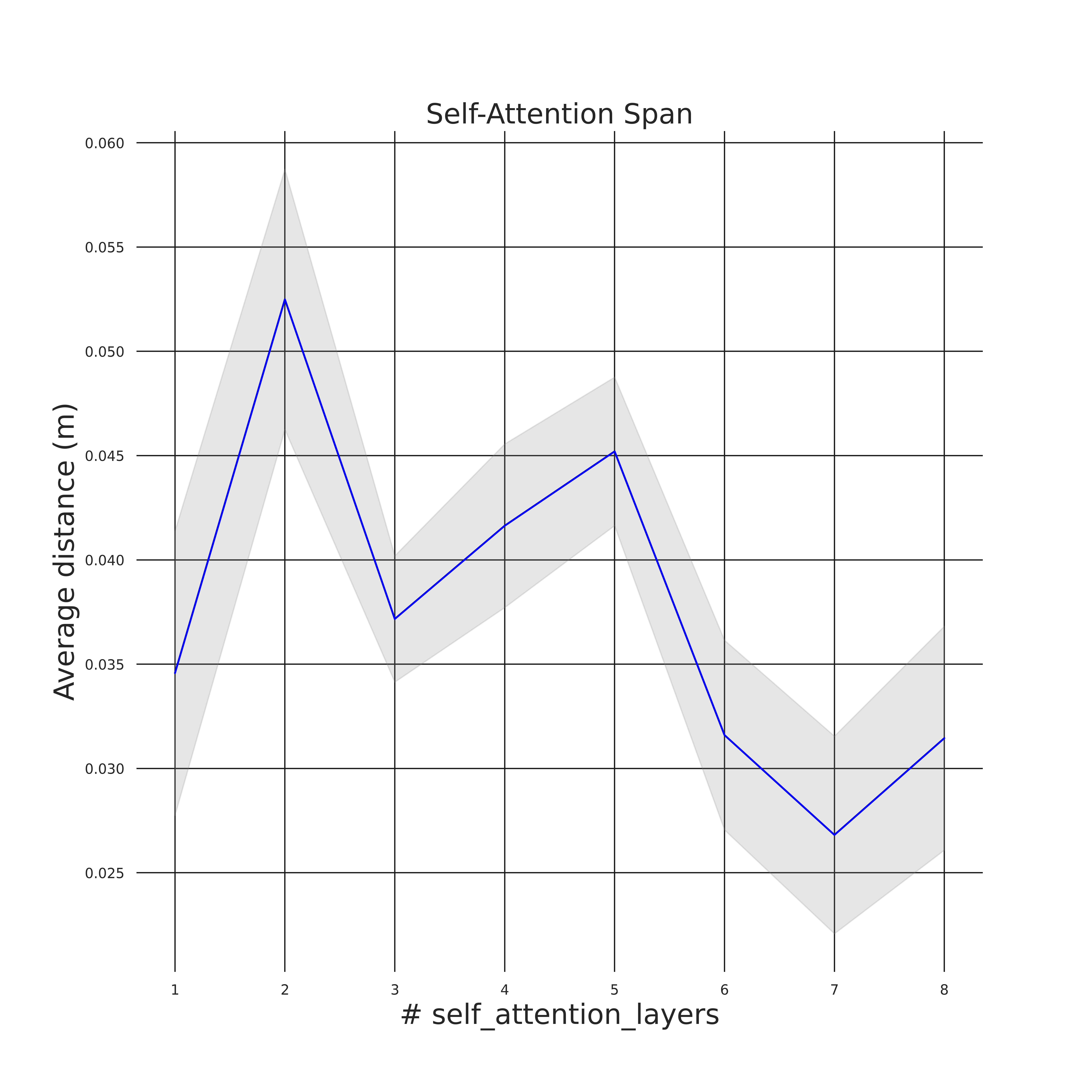}
	\vspace{-0.5em}
	\caption{
			Attention span as a function of layer depth in meters. The grey area indicates $95\%$ confidence interval.}
	\label{fig:self_attention_span}
\end{figure}

\begin{figure*}[t]
	\centering
		\includegraphics[height=0.95\textheight, width=\linewidth]{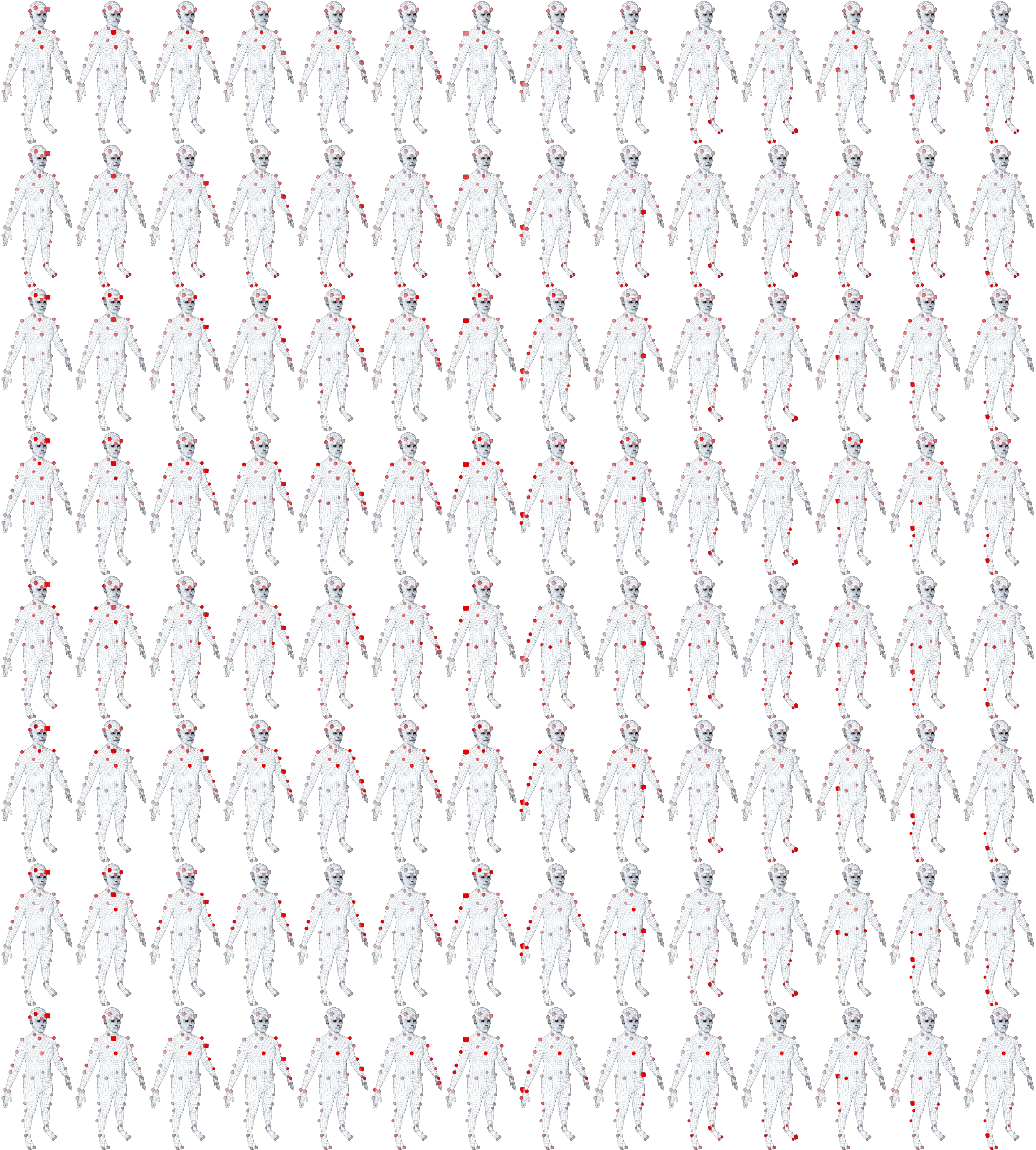}
	\vspace{-0.5em}
	\caption{
		Attention span for 14 markers, across all layers. Each row corresponds to a layer in ascending order, with bottom most row showing the last layer.}
	\label{fig:self_attention_span_extra_markers}
\end{figure*}
As explained in Sec.~\ref{sec:method_self_attention}, to increase the capacity of the network and learn rich point features at multiple levels of abstraction, we stack multiple self-attention residual layers.
Following \cite{vaswaniAttentionAllYou2017}, a transformer self-attention layer, Fig.~\ref{fig:sc_architecture}, takes as input two vectors, the query (Q), and the key (K), and computes a weight vector $ W \in [0, 1]$ that learns to focus on different regions of the input value (V), to produce the final output. 
In self-attention, all the three vectors (key, query and value) are projections of the same input; i.e.\ either 3D points or their features in deeper layers. 
All the projection operations are done by 1D-convolutions, therefore the input and the output features only differ in the last dimensions (number of channels).
Following notation of \cite{vaswaniAttentionAllYou2017}:
\begin{equation}
 	\mathrm{Attention}(Q, K, V) = \mathrm{softmax}(\frac{QK^T}{\sqrt{d_{model}}})V.
 	\label{eqn:attention}
\end{equation}
In a controlled experiment on the validation dataset, \validationDataset with marker layout presented in Fig.~\ref{fig:mlayout_hdm05}, we pass the original markers (without noise) through the network and keep track of the attention weights at each layer; i.e.\ output after Softmax in Eqn.~\ref{eqn:attention}.
At each layer, the tensor shape for the attention weights is $\#batch \times \#heads \times \#points \times \#points$. 
We concatenate frames of 50 randomly selected sequences, roughly 50000 frames, %
and take the maximum weight across heads and the mean over all the frames to arrive at a mean attention weight per layer; ($\#points \times \#points$). 
In Fig.~\ref{fig:layer_label_attention_body}, the weights are visualized on the body with a color red intensity for 3 markers. 
In the first layers, the attention span is wide and covers the entire body.
In deeper layers, the attention becomes gradually more focused on the marker of the interest and its neighboring markers on the body surface.
Fig. \ref{fig:self_attention_span_extra_markers} shows the attention span for more markers.

To make this observation more concrete, we compute the euclidean distance of each marker to all others on a A-Posed body to create a distance discrepancy matrix of ($\#points \times \#points$), and multiply the previous mean attention weights with this distance discrepancy matrix to arrive at a scalar for attention span in meters. %
On average we observe a narrower focus for all markers in deeper layers; Fig.~\ref{fig:self_attention_span}. 

\section{Implementation Details}\label{supmat:implementation_details}
\begin{figure*}
	\centering
\begin{minipage}[b]{0.3\linewidth}
		\centering
		\includegraphics[width=.8\linewidth]{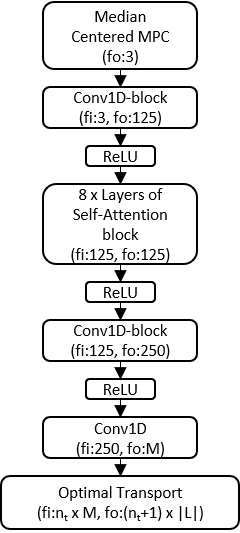}
		\subcaption{Detailed \soma Architecture}
		\label{fig:model_architecture_detailed_all}
\end{minipage}
\quad
\begin{minipage}[b]{0.3\linewidth}
		\centering
		\includegraphics[width=.7\linewidth]{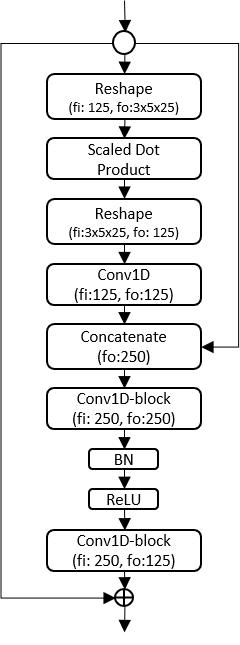}
		\subcaption{Self-Attention block}
		\label{fig:model_architecture_detailed_self_attention}
\end{minipage}
\quad
\begin{minipage}[b]{0.3\linewidth}
	\centering
	\includegraphics[width=\linewidth]{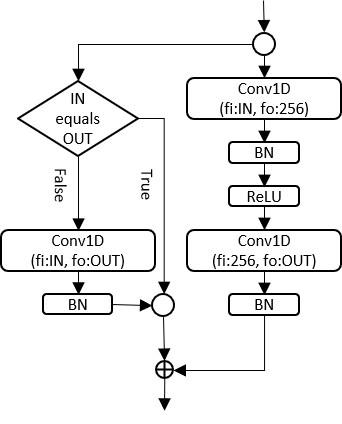}
	\subcaption{Conv1D-block}
	\label{fig:model_architecture_detailed_conv1d_block}
\end{minipage}
	\caption{
		Detailed components of \soma model. fi and fo show the number of input and output features of the layer. 
		$n_t$ is the number of points in a frame of data and $|L|$ is number of all labels including \nan. 
		IN and OUT in (c) show the number of input and output features of the block.
		All convolutions are one dimensional. BN stands for batch normalization  \cite{batchNorm2015}.
	}
	\label{fig:model_architecture_detailed}
\end{figure*}
Through model selection, Sec. \ref{sec:model_selection}, we choose $35$ iterations for Sinkhorn and $\numAttLayers=8$ as optimal choices and we empirically pick $c_{l} = 1, c_{reg}=\weightDecay, d_{model}=125, \numAttHeads=5$.
The model contains \numSOMAParameters parameters and full training on 8 Titan V100 GPUs takes roughly \trainTime hours.
We implement \soma in PyTorch \cite{pytorch_panze_2019}. %
We benefit from the log-domain stable implementation of Sinkhorn released by \cite{SuperGlue_Sarlin_2020_CVPR}.
We use ADAM \cite{kingmaba2014} with a base learning rate of \learningRate and reduce it by a factor of $0.1$ when validation error plateaus with patience of \lrSchedulerPatience epochs and train until validation error does not drop anymore for \earlyStopPatience epochs.
The training code is implemented in PyTorch Lightning \cite{falcon2019pytorch} and easily extendable to run on multiple GPUs. 
For the LogSoftmax experiment, we replace the optimal transport layer and everything else in the architecture remains the same. 
In this case, the score matrix, $S$ in Fig.~\ref{fig:sc_architecture}, will have an extra dimension for the null label. 
Fig. \ref{fig:model_architecture_detailed} shows a detailed architecture of the \soma model.

\section{Hyper-parameter Search}\label{sec:model_selection}
\begin{figure}[h]
	\centering
	\includegraphics[width=\columnwidth]{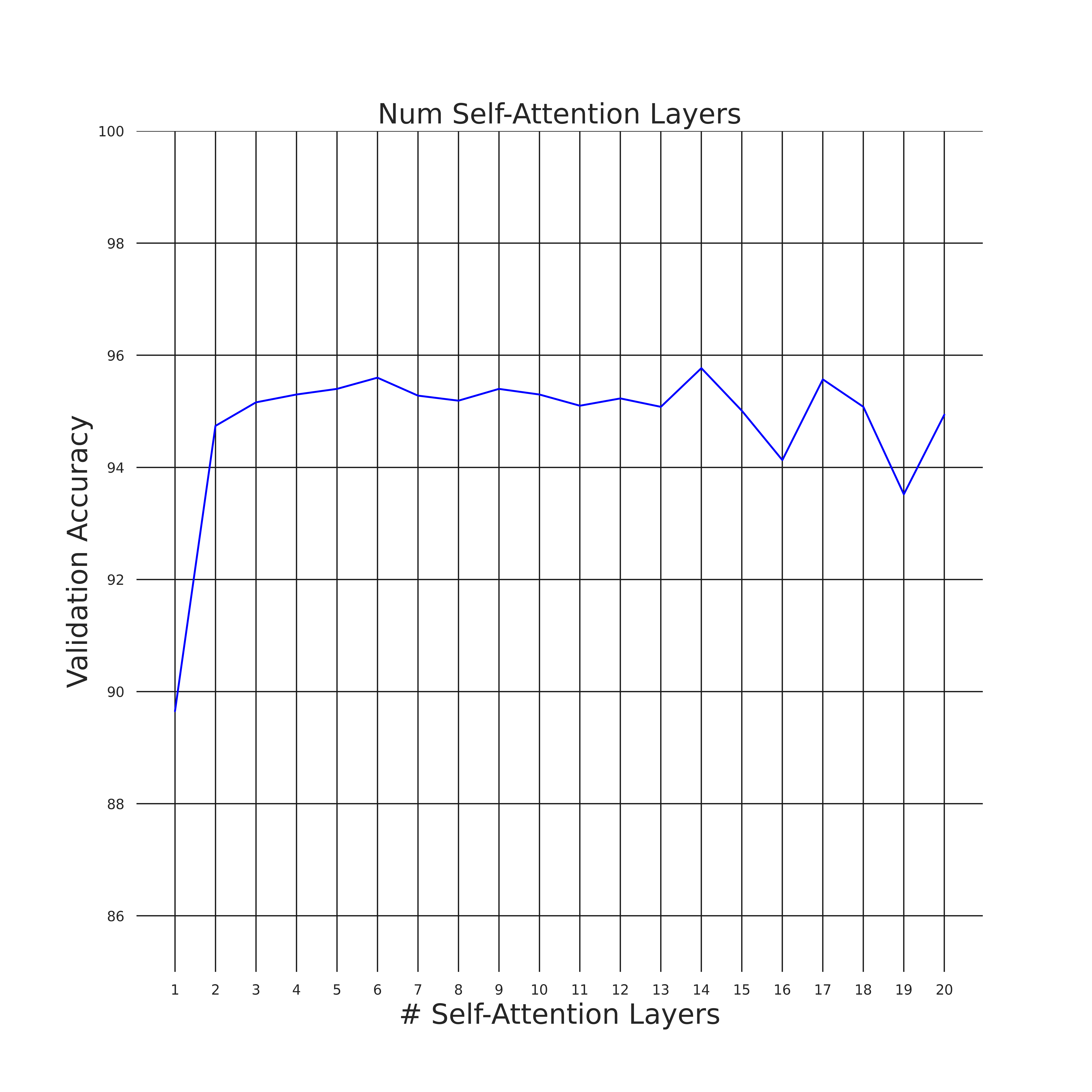}
	\caption{Validation accuracy as a function of number of attention layers}
	\label{fig:num_att_layers}
\end{figure}

\begin{figure}[h]
	\centering
	\includegraphics[width=\columnwidth]{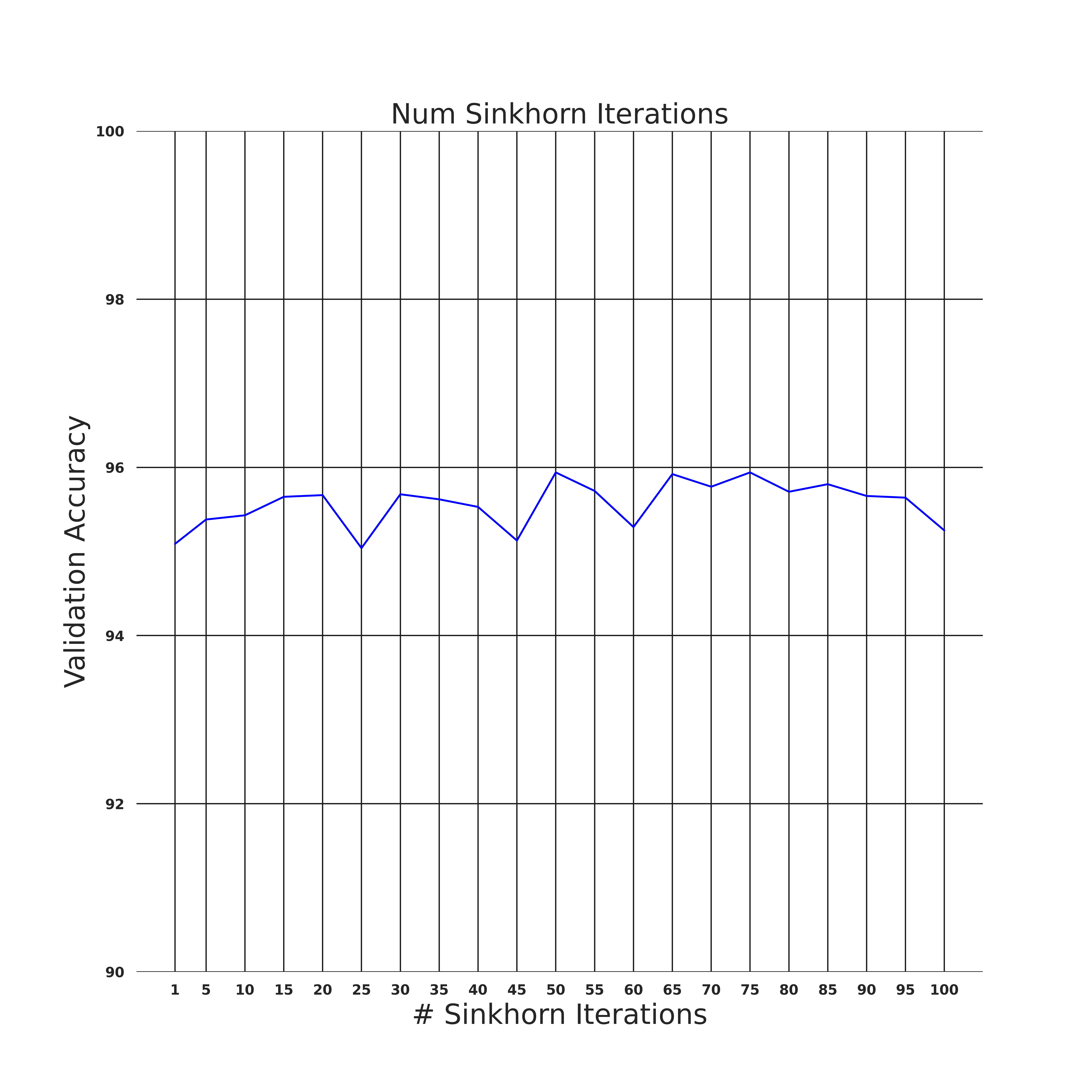}
	\caption{Validation accuracy as a function of number of Sinkhorn normalization steps.}
	\label{fig:num_sinkhorn_iters}
\end{figure}
\begin{figure}[]
	\begin{center}
		\includegraphics[width=\columnwidth]{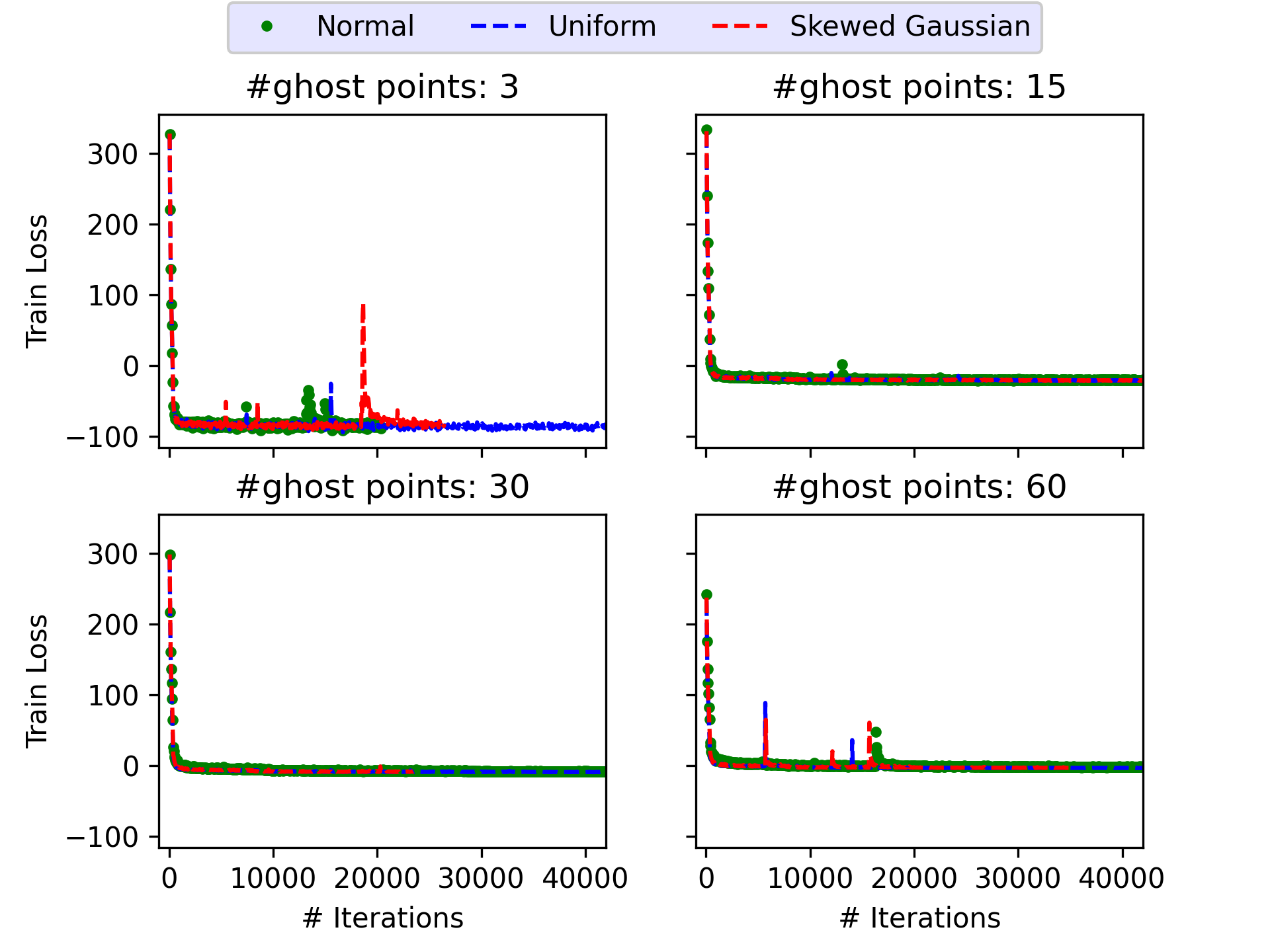}
		\caption{Training convergence with extreme
			ghost point distributions on the validation dataset for 40 training epochs; i.e. \validationDataset.}
		\label{fig:extreme_noise_train}
	\end{center}
\end{figure}
\begin{figure}
	\centering
		\centering
		\includegraphics[width=\columnwidth]{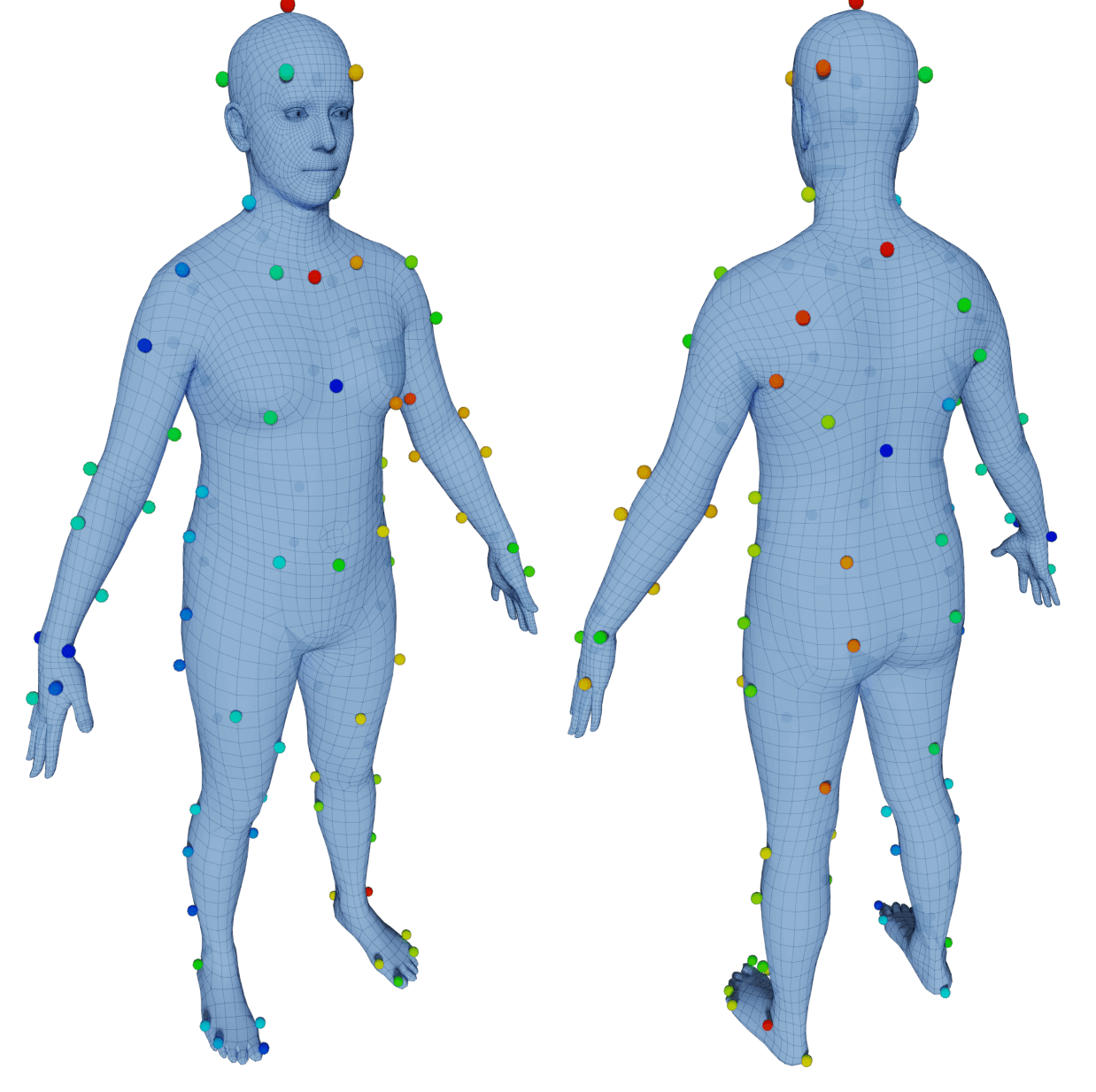}
	\caption{Marker layout from MoSh \cite{Loper:SIGASIA:2014} dataset with 89 markers. A model trained on this marker layout is used for rapid automatic label priming for labeling the single frame per significant marker layout variation.}
	\label{fig:superset_mlayout}
\end{figure}
To choose the optimum number of attention layers and iterations for Sinkhorn normalization we exploit the validation dataset \validationDataset to perform a model selection experiment. 
We produce synthetic training data following the prescription of Sec.~\ref{sec:method_synthetic_data} using the marker layout of \valds (Fig.~\ref{fig:mlayout_hdm05}) and evaluate on real markers with synthetic noise as explained in Sec.~\ref{sec:experiments}.
For hyperparameter evaluation, we want to eliminate random variations in the network weight initialization so we always use the same seed.
In Fig.~\ref{fig:num_att_layers}, we train one model per given number of layers. 
Guided by this graph we choose $\numAttLayers=\numAttentionLayers$ layers as a sensible choice for adequate model capacity, i.e.\ $\numSOMAParameters$, and generalization to real markers.
In Fig.~\ref{fig:num_sinkhorn_iters}, we repeat the same process, this time keeping the number of layers fixed as $\numAttentionLayers$, and varying the number of Sinkhorn iterations. 
We choose $\numSinkhornIter$ iterations that seem a good trade-off between computation time vs performance. %

\section{Standard Deviations}
\begin{table}[]
	\centering
	\resizebox{\columnwidth}{!}{%
		\begin{tabular}{lccccccc}
			\hline
			\multicolumn{1}{c}{\multirow{2}{*}{Method}} & \multicolumn{7}{c}{Number of Exact Per-Frame Occlusions} \\ \cline{2-8} 
			\multicolumn{1}{c}{}     & 0    & 1    & 2    & 3    & 4    & 5    & 5+G  \\ \hline
			\textbf{\soma-Real}      & 1.76 & 1.90 & 2.03 & 2.22 & 2.44 & 2.73 & 3.08 \\
			\textbf{\soma-Synthetic} & 4.68 & 2.89 & 3.25 & 3.54 & 4.10 & 4.62 & 5.92 \\
			\textbf{\soma *}         & 1.59 & 1.76 & 1.91 & 2.12 & 2.34 & 2.59 & 2.85 \\ \hline
		\end{tabular}%
	}
	\caption{Accuracy standard deviation corresponding to Tab. 2 of the main paper.}
	\label{tab:comparision_previous_work_std}
\end{table}
In Tab.~\ref{tab:comparision_previous_work_std}, we report accuracy standard deviation of Tab.~\ref{tab:comparision_previous_work} as complementary material. We observe lower variation for the model trained on synthetic data using \amass bodies. The model trained on synthetic data of limited number of bodies %
 shows the largest variation.
\section{Marker Layout Variation of \validationDataset}
\begin{figure*}
	\centering
	\begin{subfigure}[b]{0.475\textwidth}   
		\centering
		\includegraphics[width=\linewidth]{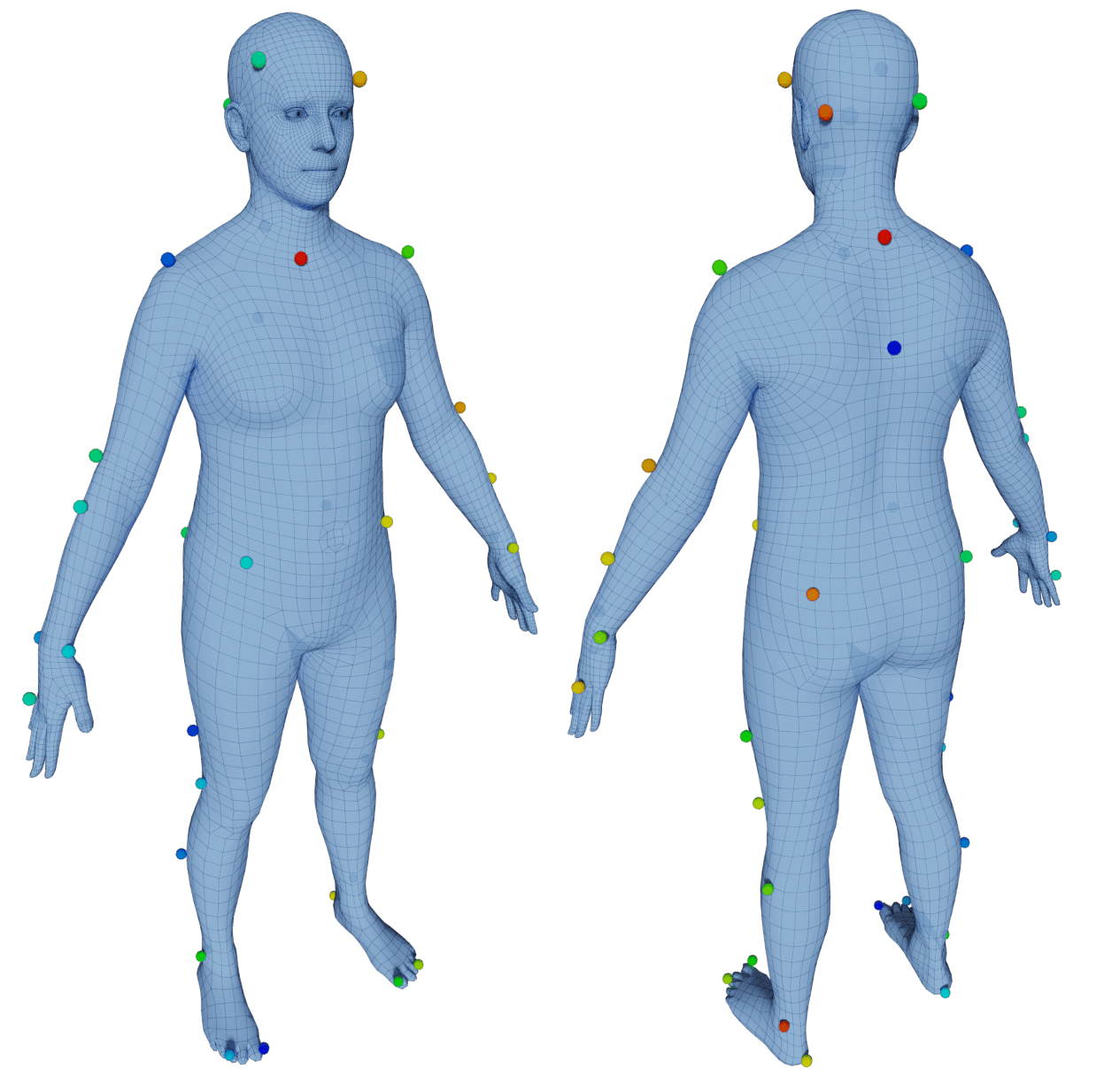}
		\caption{}
		\label{fig:hdm05_3}
	\end{subfigure}%
	\hfill
	\begin{subfigure}[b]{0.475\textwidth} 
		\centering
		\includegraphics[width=\linewidth]{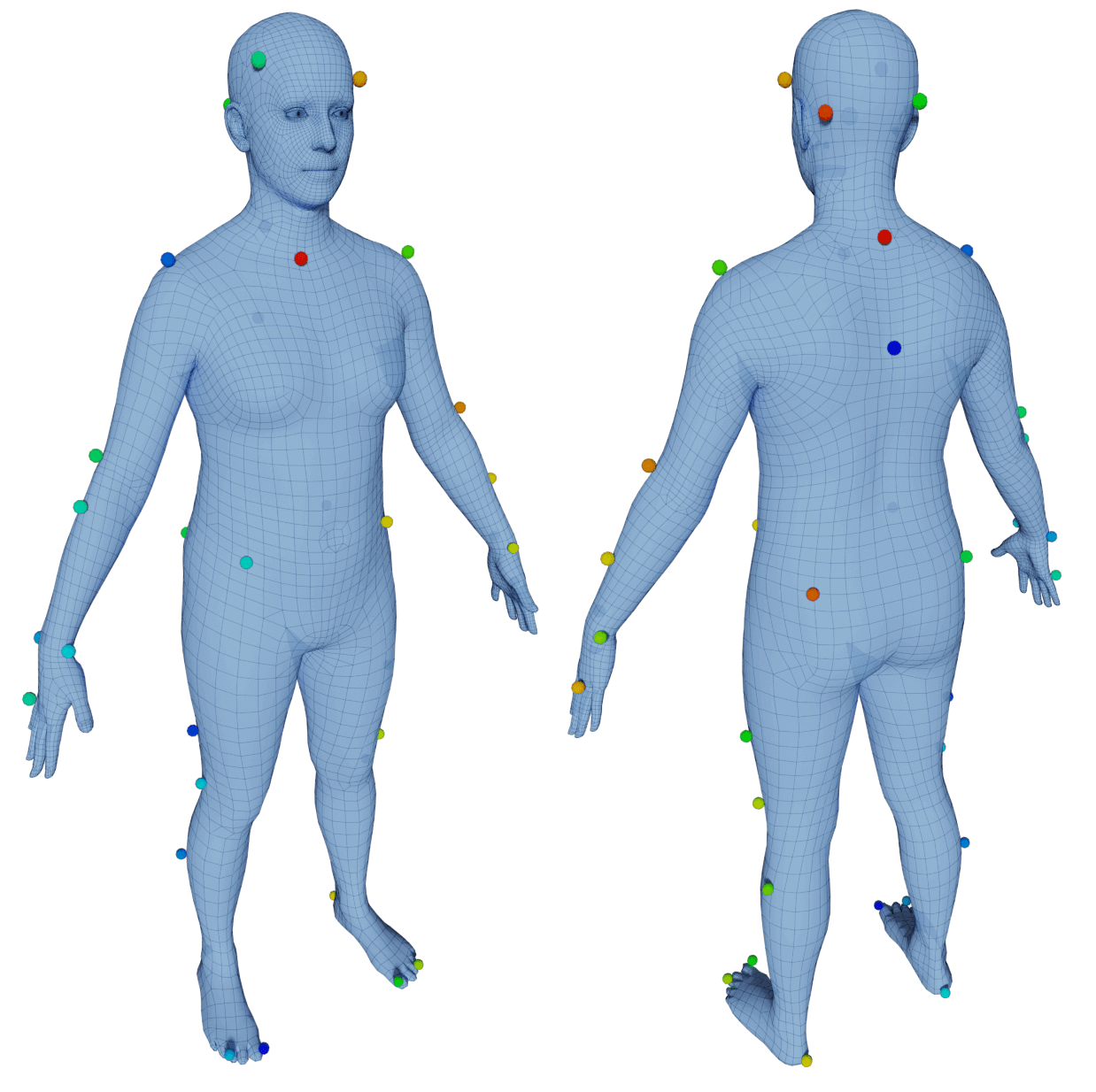}
		\caption{}
		\label{fig:hdm05_5}
	\end{subfigure}
	\vskip\baselineskip
	\begin{subfigure}[b]{0.475\textwidth}
		\centering
		\includegraphics[width=\linewidth]{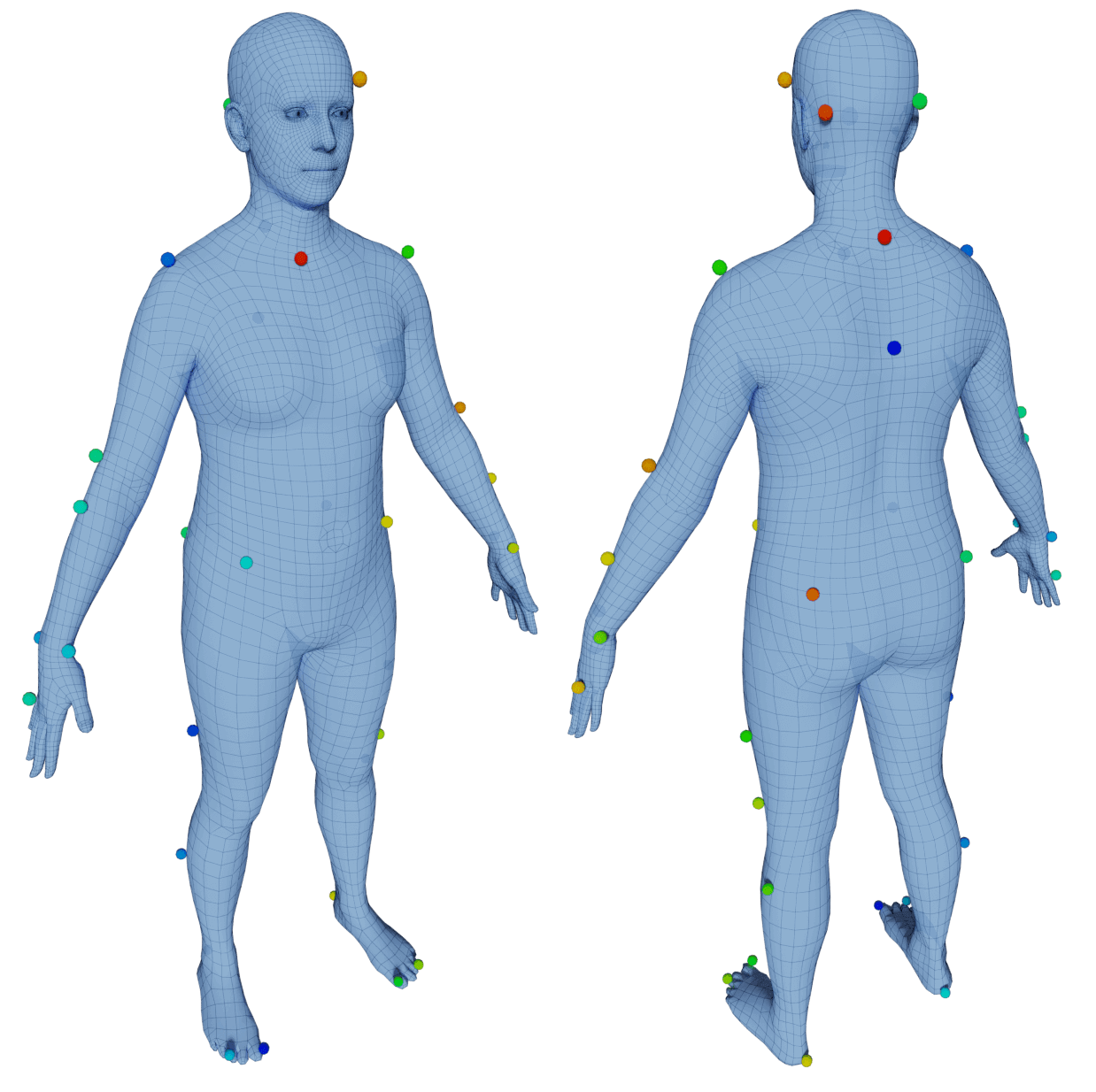}
		\caption{}
		\label{fig:hdm05_6}
	\end{subfigure}
	\hfill
	\begin{subfigure}[b]{0.475\textwidth}
		\centering
		\includegraphics[width=\linewidth]{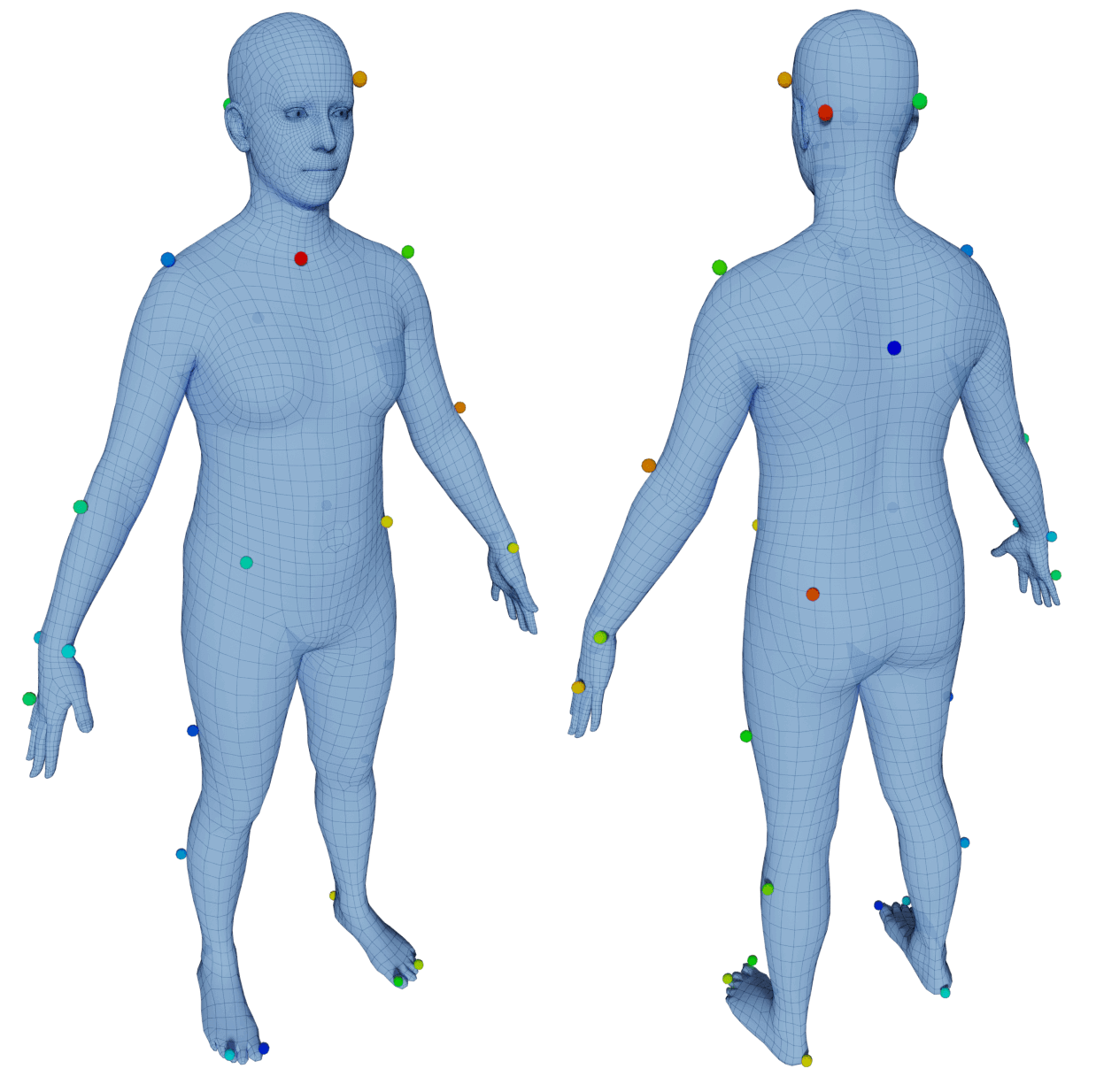}
		\caption{}
		\label{fig:hdm05_12}
	\end{subfigure}
	\caption{Modified \valds marker layout. %
		Number of markers removed: (a) 3 (b) 5 (c) 9 (d) 12.}
	\label{fig:hdm05_variations}
\end{figure*}
In Fig.~\ref{fig:hdm05_variations} we visualize the marker layout modifications for the experiment in Sec.~\ref{sec:experiments_mocap_setup_variation}.

\section{Stability of the Training Process}\label{sec:training_stability}

We consistently observe stable runtime and training processes.
In Fig.~\ref{fig:extreme_noise_train}, we provide training
curves for more ``extreme" ghost point distributions. %
Specifically, we add a uniform distribution in a cubic volume in the range of $[-2,2]$ meters and skewed Gaussian with a
mean location sampled uniformly from the same random volume and a random covariance matrix. %
We also drastically increase the number of ghost points to up to 60 per-frame.
As suggested by the figure, training is stable and converges from early iterations on.

\section{Processing Real \Mocap Data}\label{supmat:processing_real_mocap}
\begin{figure*}
	\centering
    \begin{subfigure}[b]{0.475\textwidth}   
		\centering
		\includegraphics[width=\linewidth]{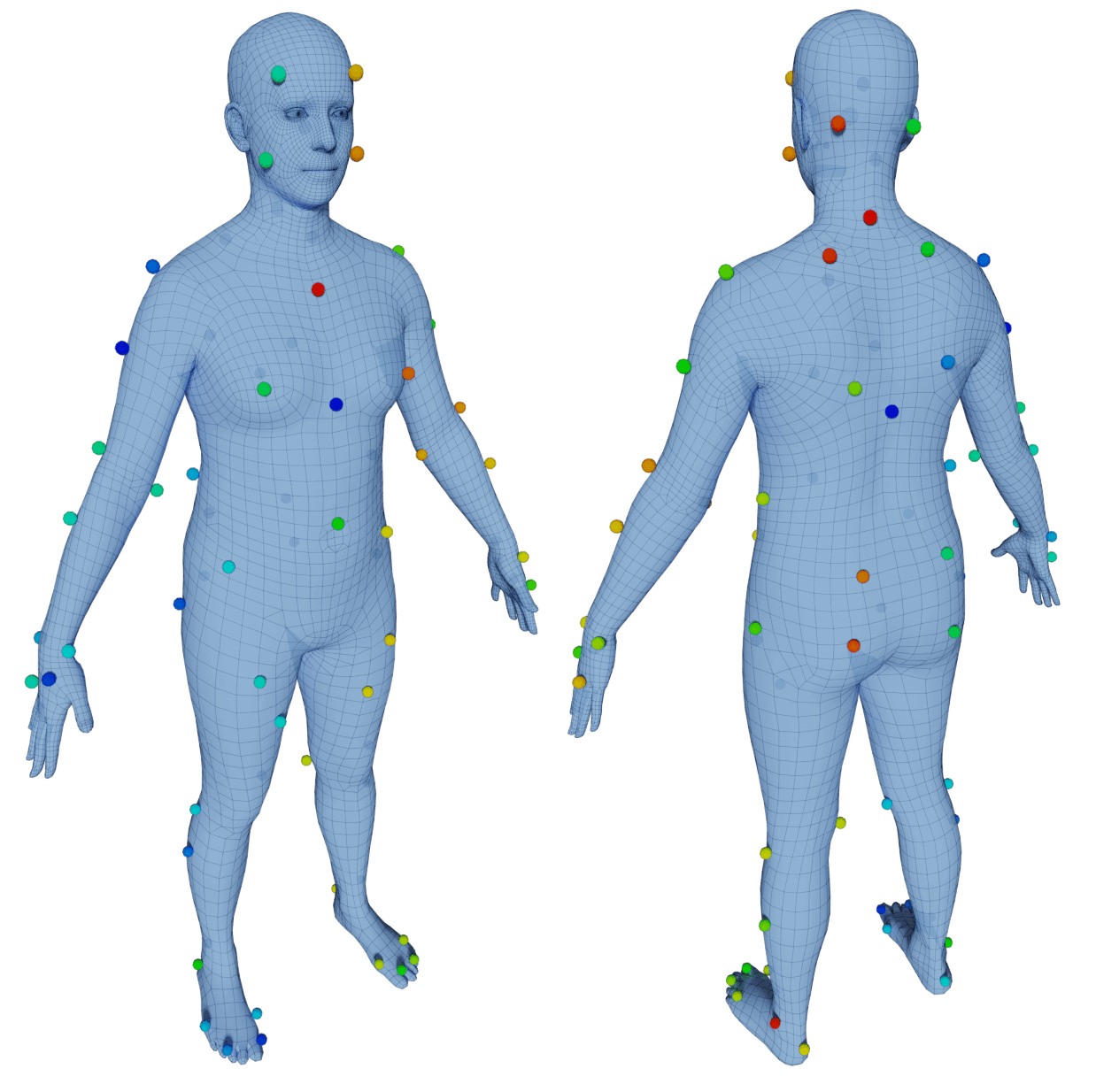}
		\caption{BMLmovi}
		\label{fig:mlayout_bmlmovi}
	\end{subfigure}%
    \hfill
	\begin{subfigure}[b]{0.475\textwidth} 
		\centering
		\includegraphics[width=\linewidth]{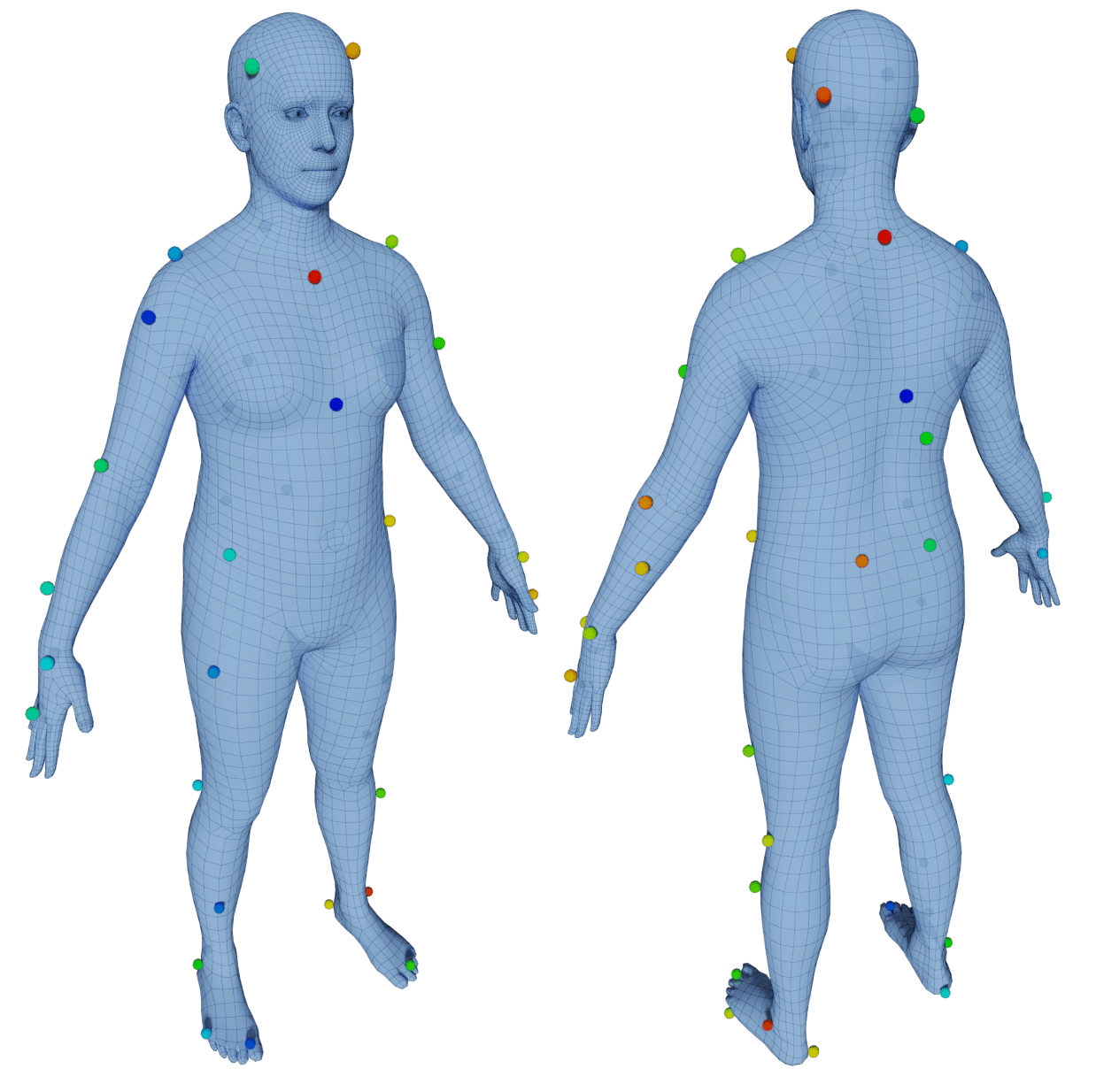}
		\caption{BMLrub}
		\label{fig:mlayout_bmlrub}
	\end{subfigure}
    \vskip\baselineskip
	\begin{subfigure}[b]{0.475\textwidth}
		\centering
		\includegraphics[width=\linewidth]{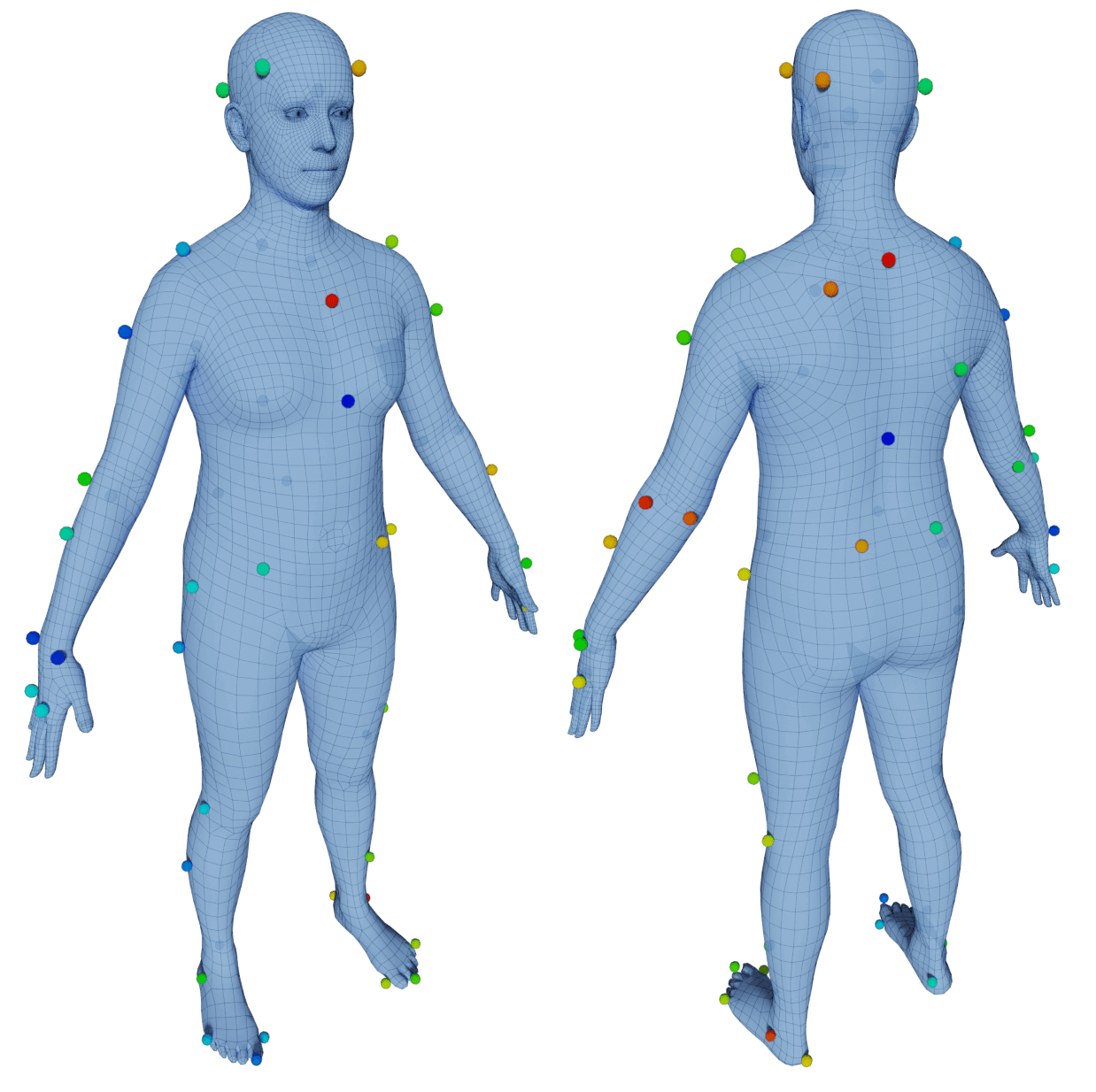}
		\caption{KIT}
		\label{fig:mlayout_kit}
	\end{subfigure}
	\hfill
	\begin{subfigure}[b]{0.475\textwidth}
		\centering
		\includegraphics[width=\linewidth]{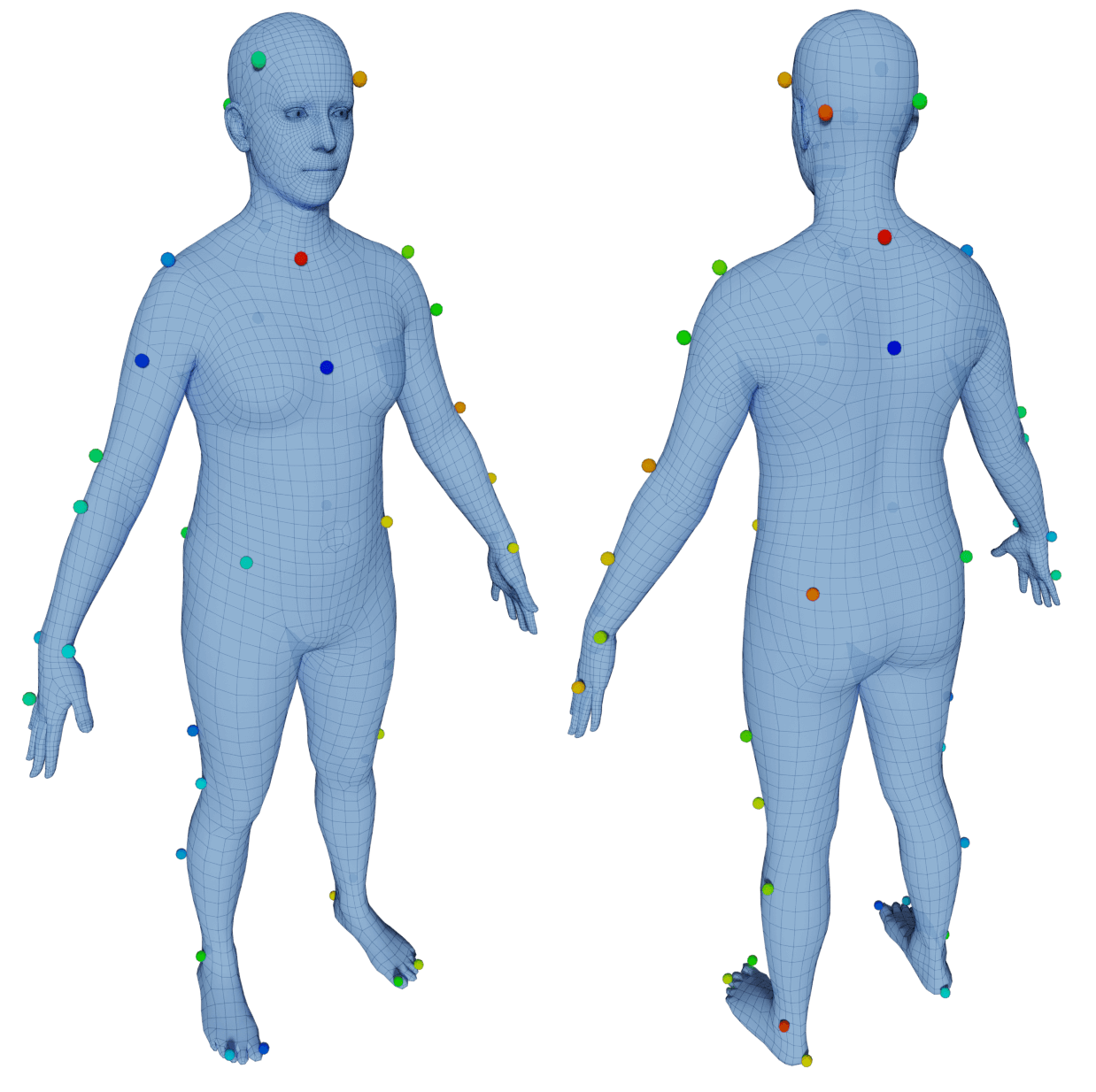}
		\caption{HDM05}
		\label{fig:mlayout_hdm05}
	\end{subfigure}
	\caption{Marker layout of test and validation datasets. %
	}
	\label{fig:testset_mlayouts}
\end{figure*}
\begin{figure*}
	\centering
	\begin{subfigure}[b]{0.45\textwidth}   
		\centering
		\includegraphics[width=\linewidth]{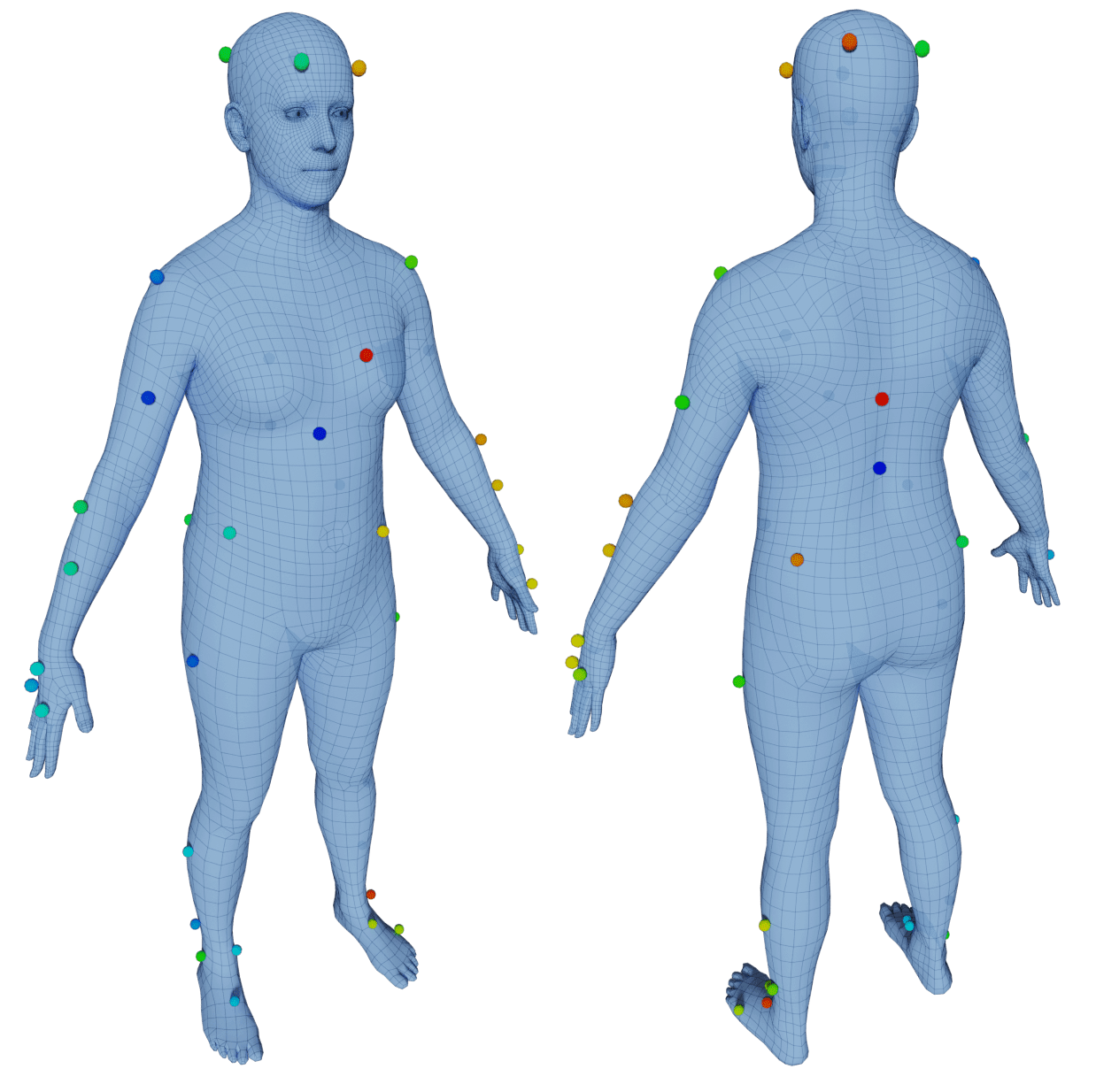}
	\end{subfigure}%
	\hfill
	\begin{subfigure}[b]{0.45\textwidth}  
		\centering
		\includegraphics[width=\linewidth]{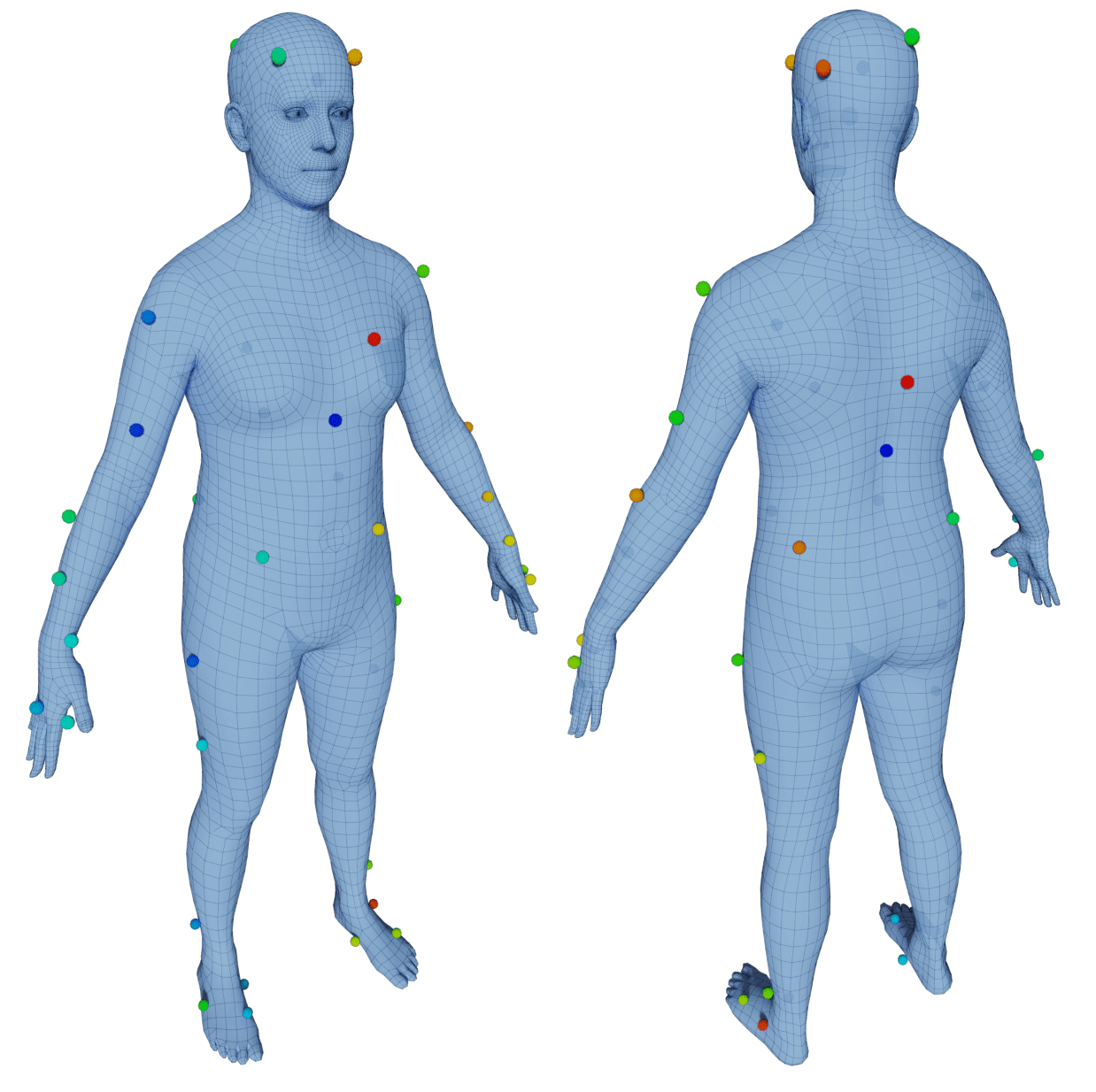}
	\end{subfigure}
	\vskip\baselineskip
	\begin{subfigure}[b]{0.45\textwidth}  
		\centering
		\includegraphics[width=\linewidth]{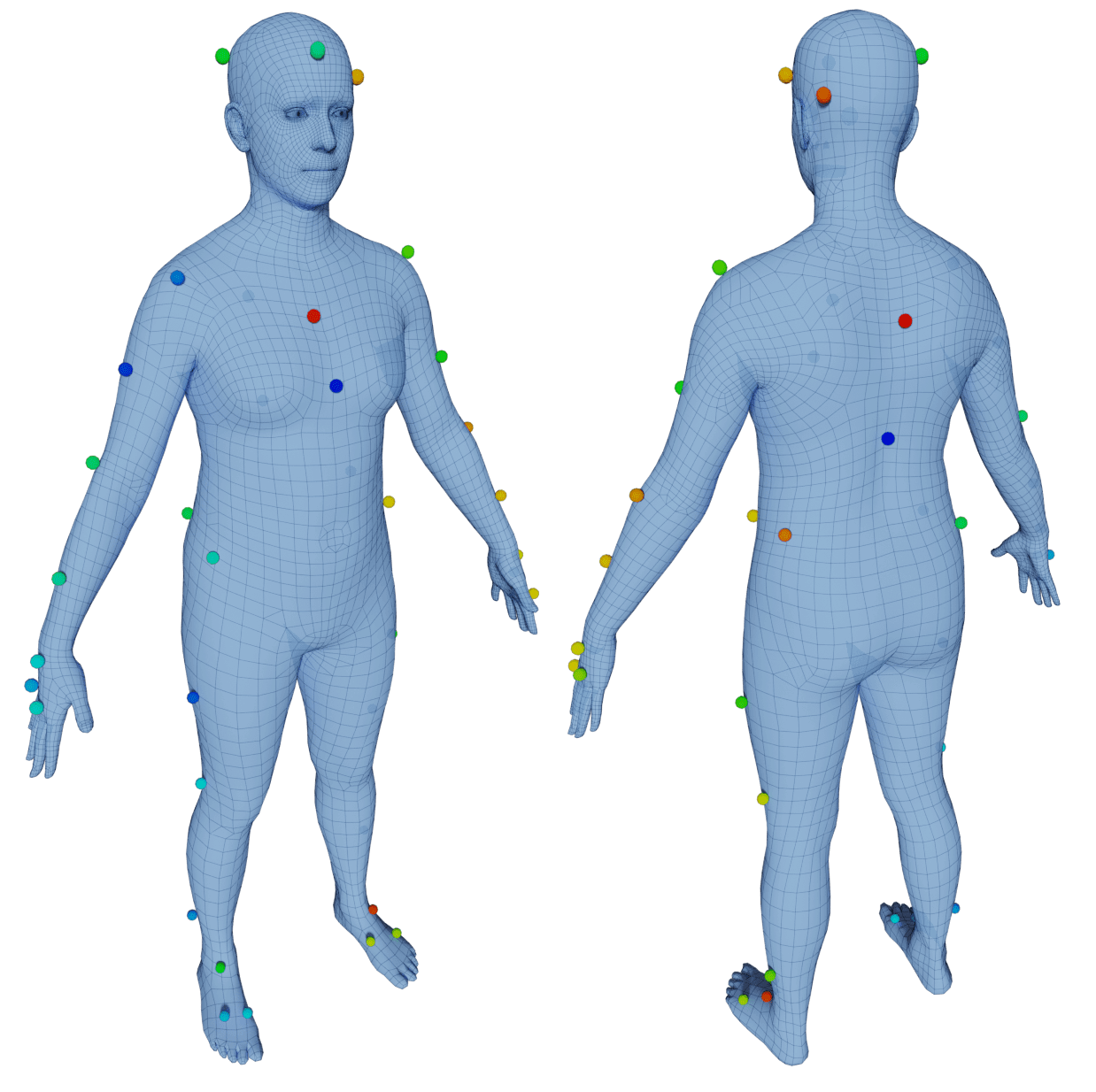}
	\end{subfigure}
	\hfill
	\begin{subfigure}[b]{0.45\textwidth}  
		\centering
		\includegraphics[width=\linewidth]{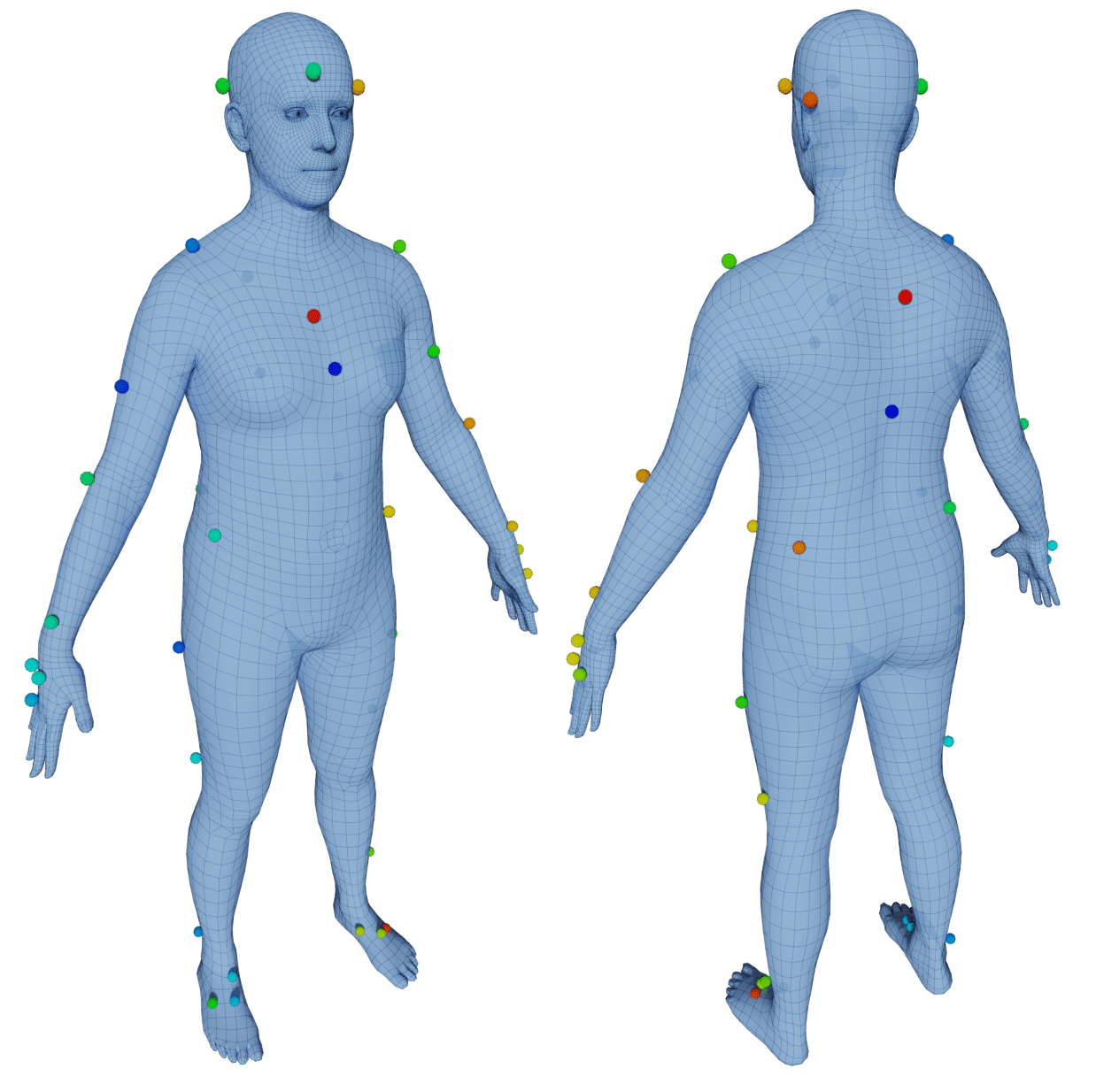}
	\end{subfigure}
	\vskip\baselineskip
	\begin{subfigure}[b]{0.45\textwidth}  
		\centering
		\includegraphics[width=\linewidth]{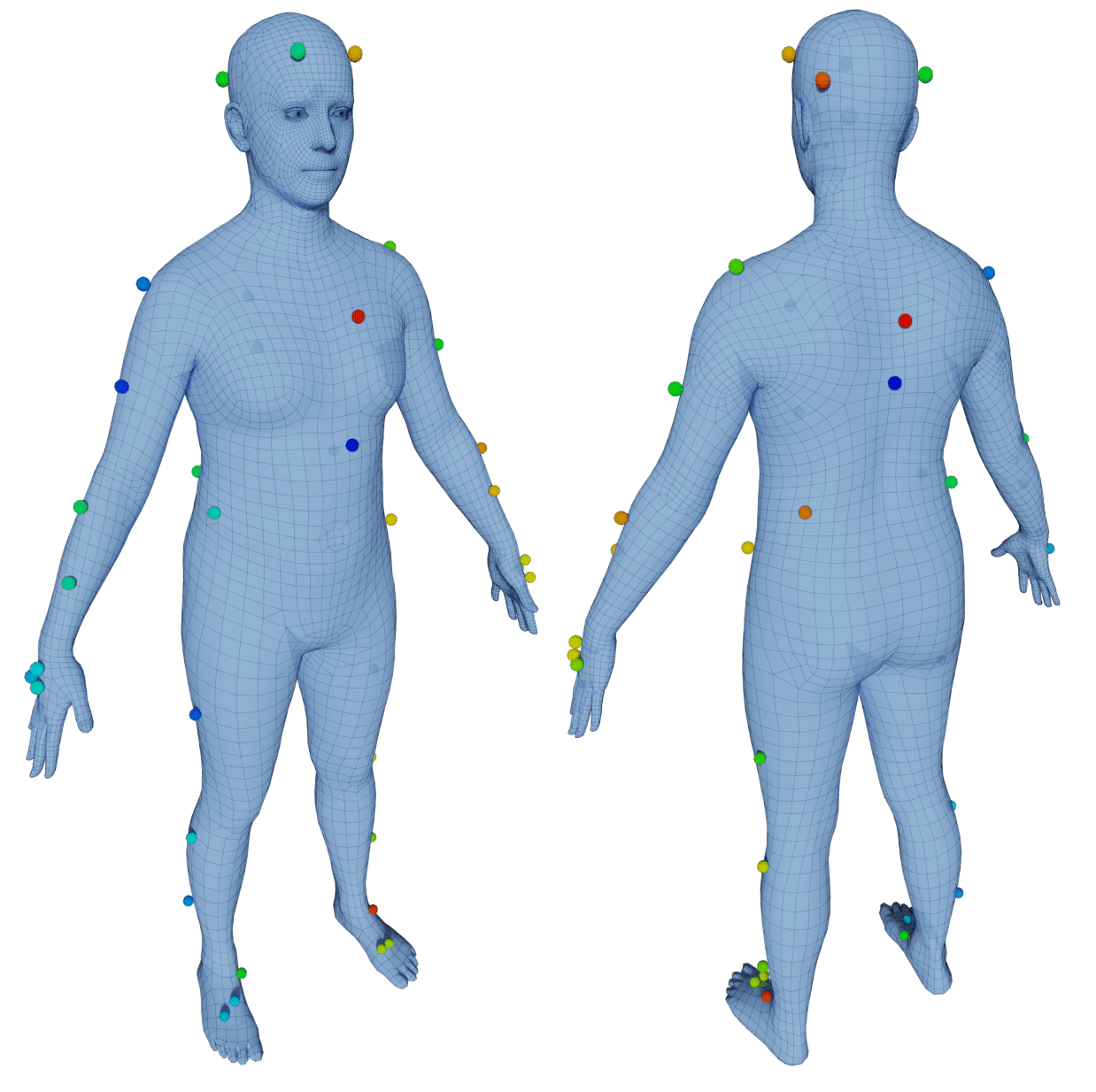}
	\end{subfigure}
	\hfill
	\begin{subfigure}[b]{0.45\textwidth}  
		\centering
		\includegraphics[width=\linewidth]{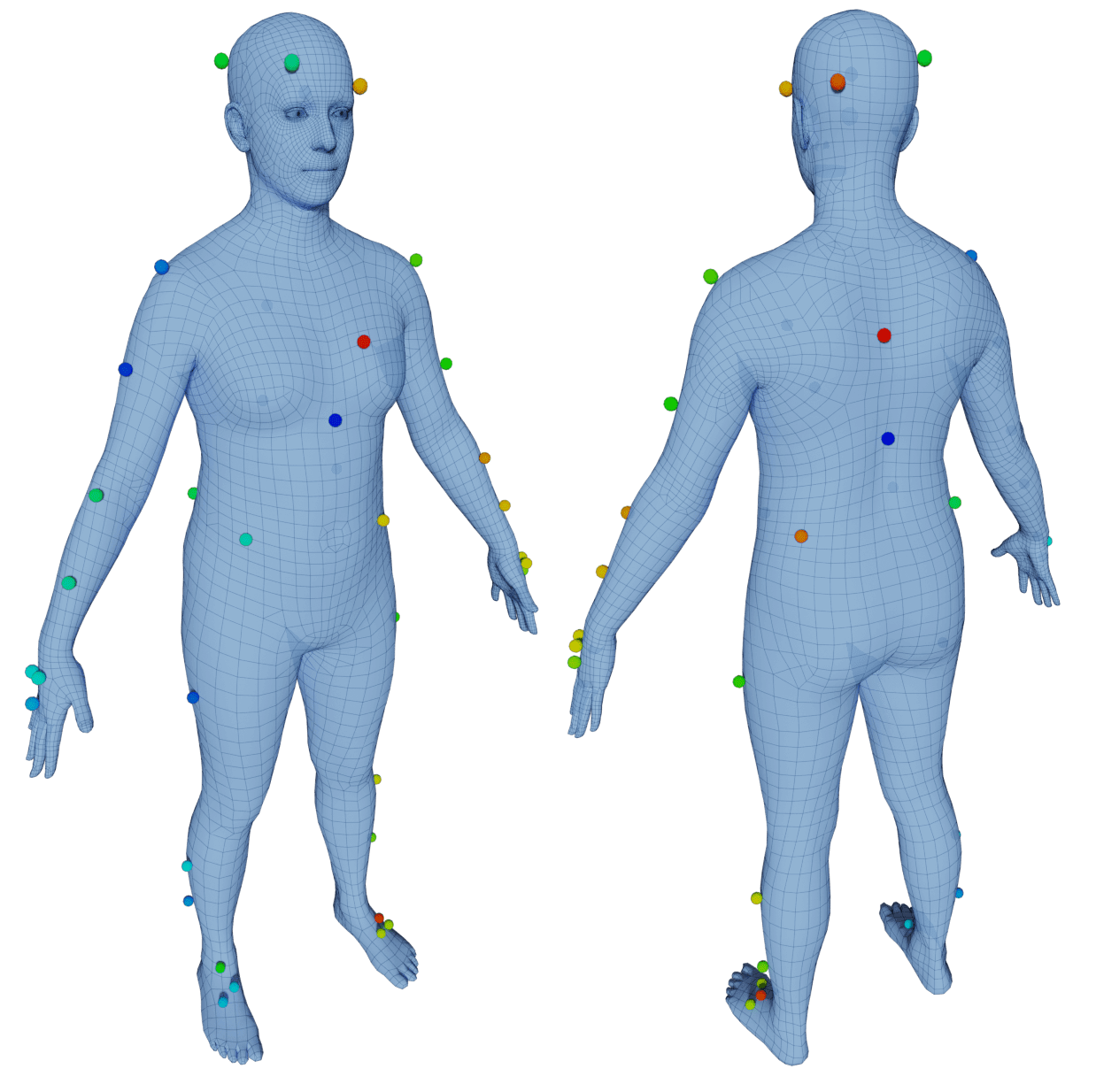}
	\end{subfigure}
	\caption{Significant variation of marker placement of DanceDB dataset on hands and foot.}
	\label{fig:dancedb_mlyouts}
\end{figure*}
\begin{figure*}
	\centering
	\begin{subfigure}[b]{0.45\textwidth}   
		\centering
		\includegraphics[width=\linewidth]{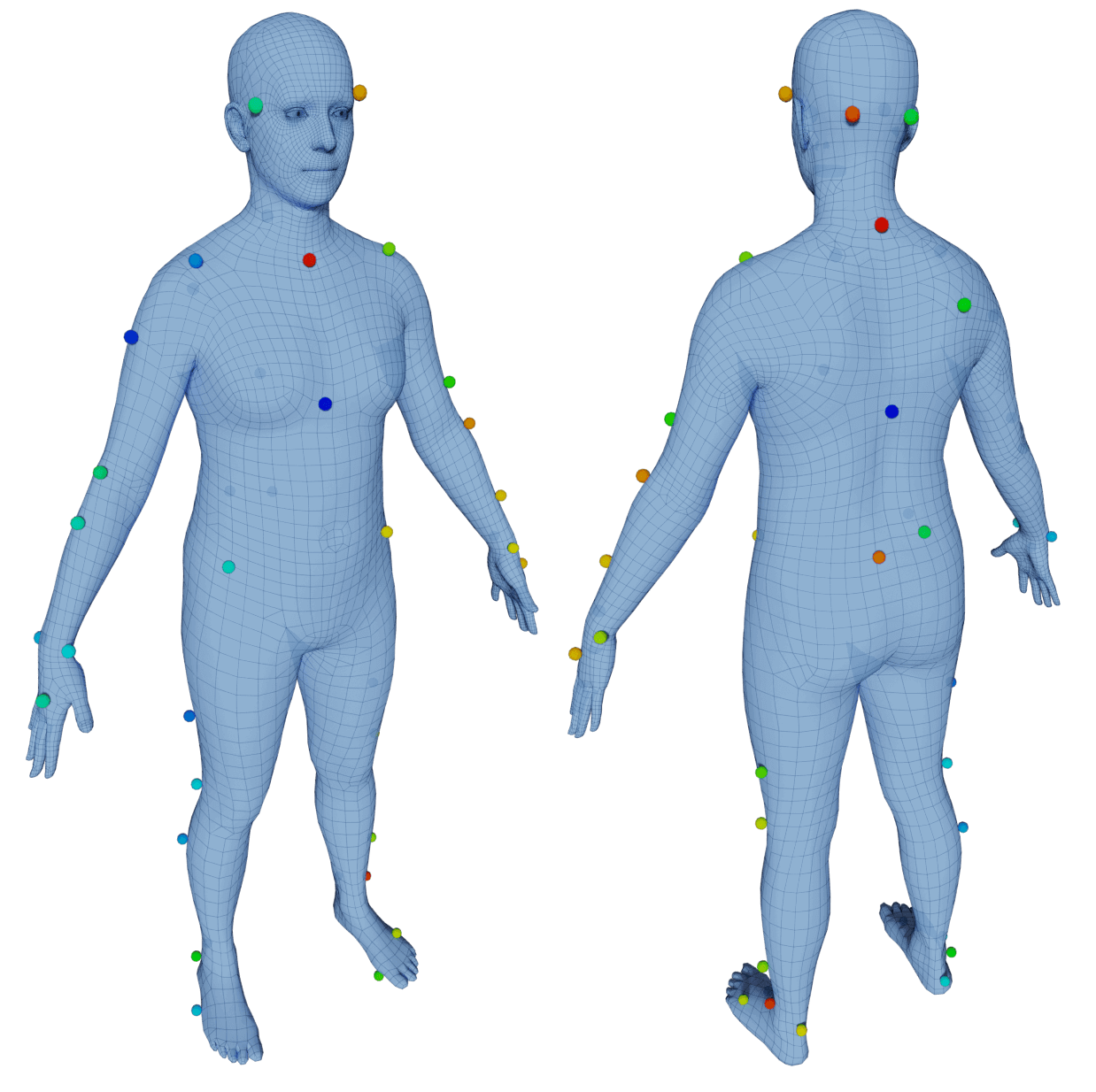}
	\end{subfigure}%
	\hfill
	\begin{subfigure}[b]{0.45\textwidth}  
		\centering
		\includegraphics[width=\linewidth]{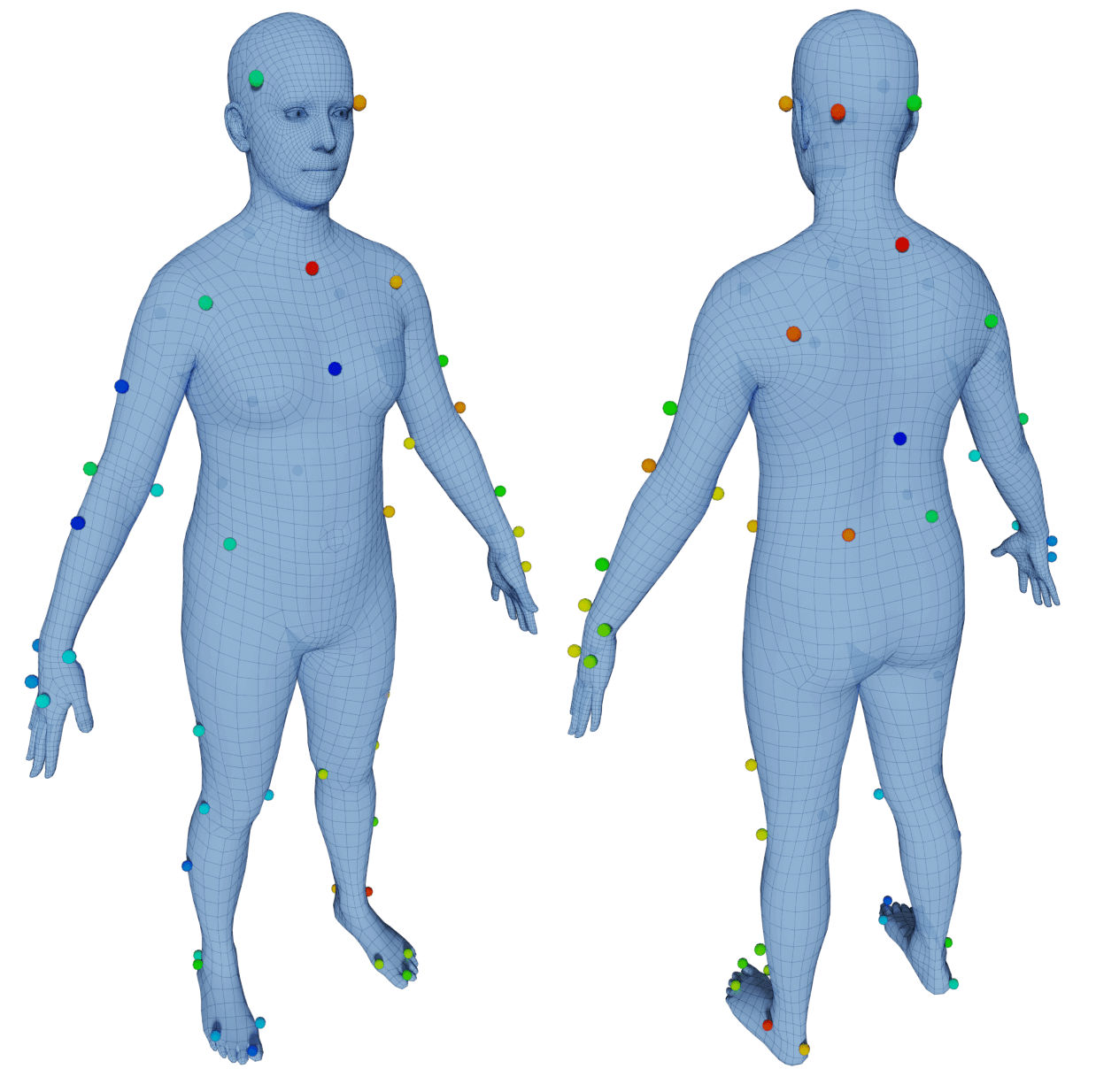}
	\end{subfigure}
	\vskip\baselineskip
	\begin{subfigure}[b]{0.45\textwidth}  
		\centering
		\includegraphics[width=\linewidth]{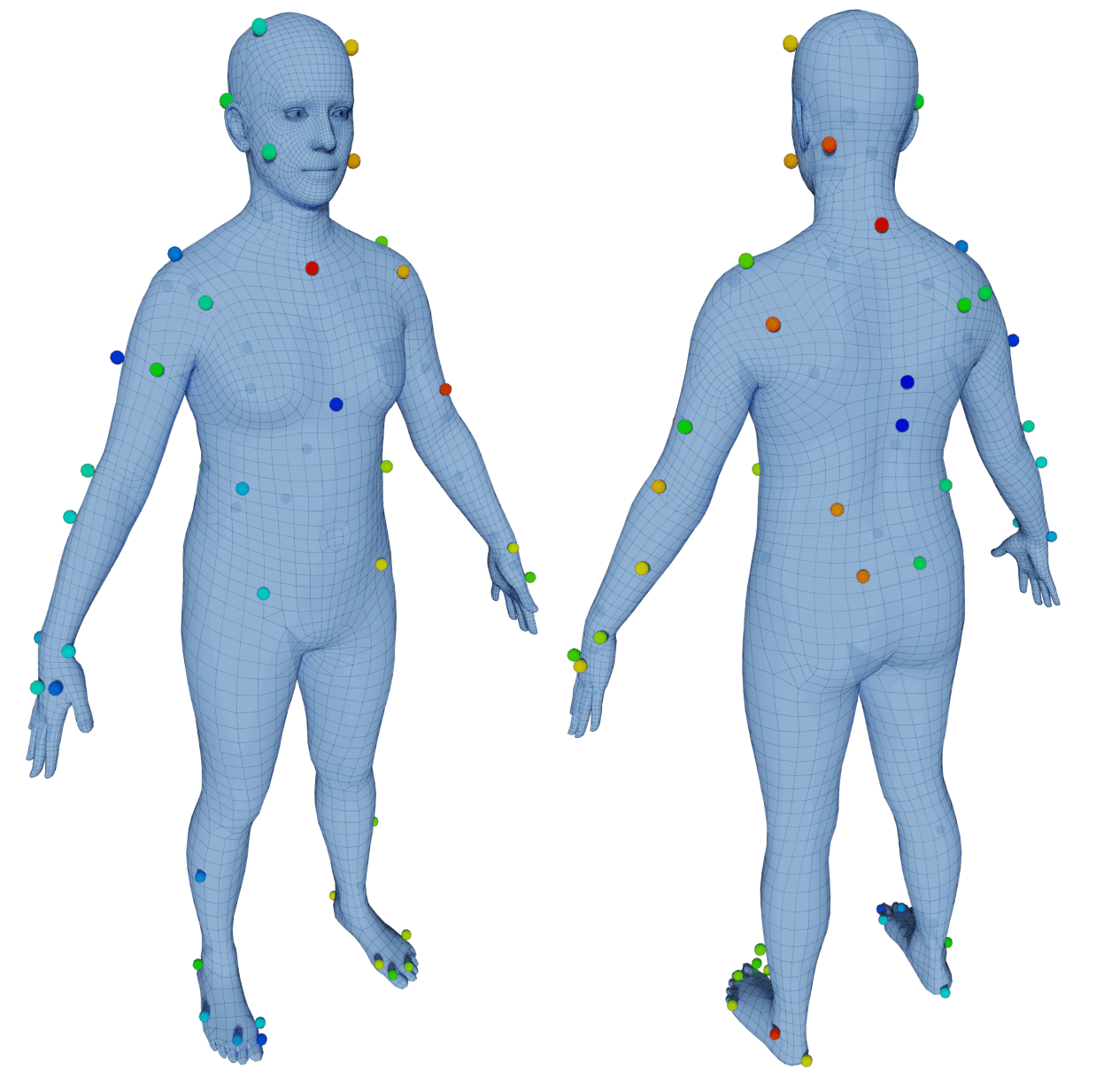}
	\end{subfigure}
	\hfill
	\begin{subfigure}[b]{0.45\textwidth}  
		\centering
		\includegraphics[width=\linewidth]{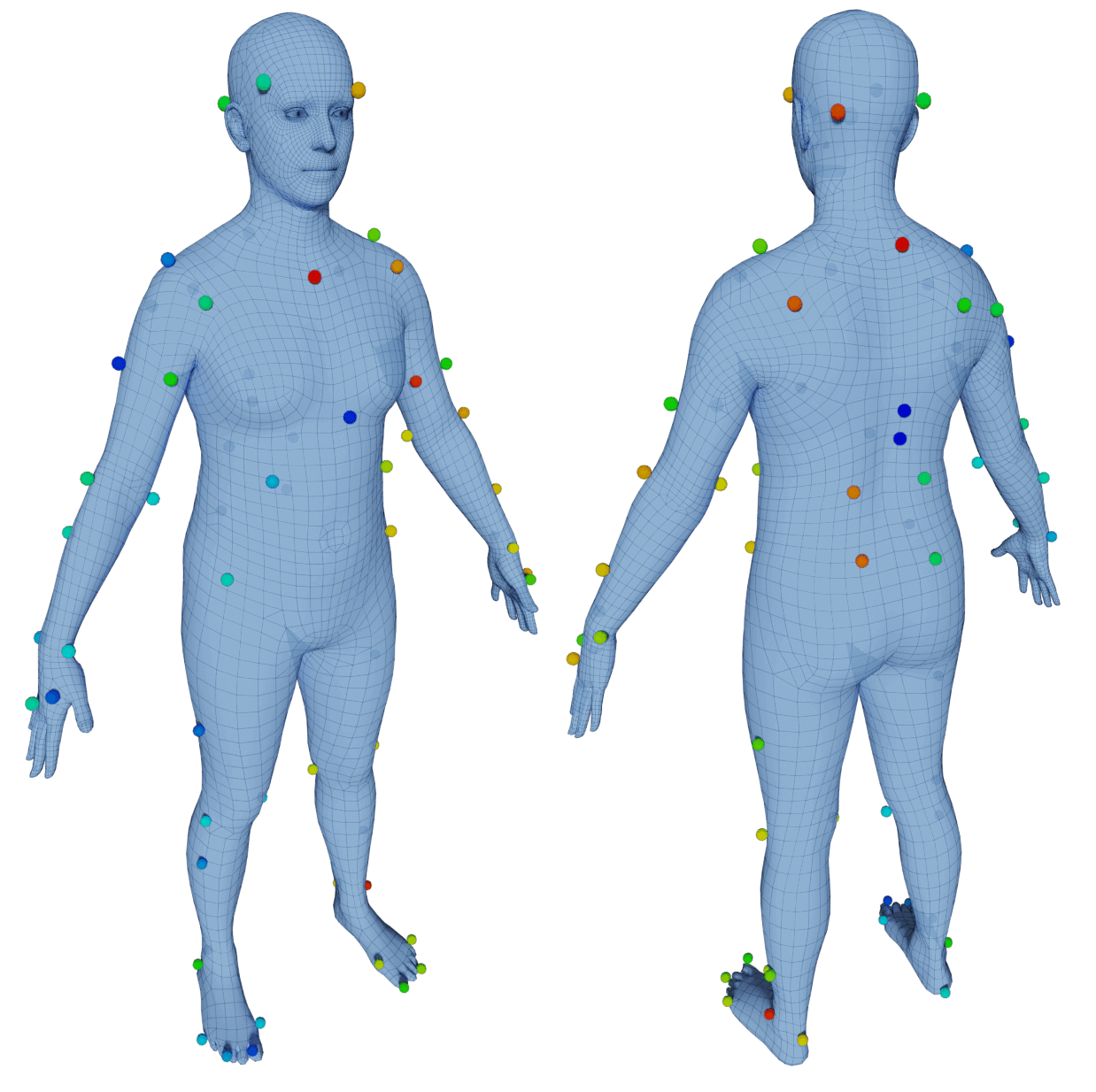}
	\end{subfigure}
	\caption{Sample of marker layouts used for training \soma model for \cmuii dataset.}
	\label{fig:cmuii_mlyouts}
\end{figure*}

\begin{table*}[]
	\centering
	\resizebox{0.8\textwidth}{!}{%
		\begin{tabular}{ccccccc}
			\hline
			Name   & $\#$ Frames & $\#$ Motions & Acc.              & F1                & $V2V_{mm}^{mean}$ & $V2V_{mm}^{median}$ \\ \hline
			Clap   & 7572        & 6            & 100.00 $\pm$ 0.00 & 100.00 $\pm$ 0.00 & 0.00 $\pm$ 0.08   & 0.00                \\
			Dance  & 15023       & 8            & 99.78 $\pm$ 0.87  & 99.68 $\pm$ 1.29  & 0.15 $\pm$ 1.76   & 0.00                \\
			Jump   & 9621        & 6            & 99.99 $\pm$ 0.13  & 99.99 $\pm$ 0.25  & 0.03 $\pm$ 0.72   & 0.00                \\
			Kick   & 10787       & 6            & 99.59 $\pm$ 1.18  & 99.48 $\pm$ 1.50  & 0.75 $\pm$ 6.92   & 0.00                \\
			Lift   & 16932       & 6            & 100.00 $\pm$ 0.00 & 100.00 $\pm$ 0.00 & 0.00 $\pm$ 0.06   & 0.00                \\
			Random & 19617       & 7            & 100.00 $\pm$ 0.05 & 100.00 $\pm$ 0.09 & 0.00 $\pm$ 0.21   & 0.00                \\
			Run    & 9356        & 6            & 100.00 $\pm$ 0.00 & 100.00 $\pm$ 0.00 & 0.00 $\pm$ 0.06   & 0.00                \\
			Sit    & 9829        & 6            & 100.00 $\pm$ 0.00 & 100.00 $\pm$ 0.00 & 0.00 $\pm$ 0.09   & 0.00                \\
			Squat  & 11287       & 6            & 100.00 $\pm$ 0.00 & 100.00 $\pm$ 0.00 & 0.01 $\pm$ 0.13   & 0.00                \\
			Throw  & 9292        & 6            & 99.99 $\pm$ 0.15  & 99.99 $\pm$ 0.22  & 0.00 $\pm$ 0.09   & 0.00                \\
			Walk   & 12264       & 6            & 100.00 $\pm$ 0.00 & 100.00 $\pm$ 0.00 & 0.00 $\pm$ 0.11   & 0.00                \\ \hline
			\rowcolor[HTML]{EFEFEF} 
			& 131580      & 69           & 99.94 $\pm$ 0.47  & 99.92 $\pm$ 0.64  & 0.08 $\pm$ 2.09   & 0.00                \\ \hline
		\end{tabular}%
	}
	\caption{Per-motion-class statistics of the \soma dataset and performance of the \soma model.}
	\label{tab:soma_dataset_details}
\vspace*{-1.2em}
\end{table*}
Here we elaborate on Sec.~\ref{sec:method_implementation}, namely on real use-case scenarios of \soma.
The marker layout of the test datasets, Sec.~\ref{sec:experiments}, are obtained by running \mosh on a single random frame chosen from the respective dataset. Fig.~\ref{fig:testset_mlayouts} demonstrates the marker layout used for training \soma for each dataset. %

In addition to test datasets with synthetic noise, presented in Sec.~\ref{sec:method_implementation}, we demonstrate the real application of \soma by automatically labeling four real \mocap datasets captured with different technologies; namely: two with passive markers, \soma and CMU-II \cite{cmuWEB}, and two with active marker technology, namely DanceDB \cite{DanceDB:2018:aristidou} and Mixamo \cite{mixamoWEB}; for an overview refer to Tab.~\ref{sec:conclusion}.

For proper training of \soma we require one labeled frame per significant variation of the marker layout throughout the dataset. Most of the time one layout is utilized to capture the entire dataset, yet as we see next, this is not always the case, especially when the marker layout is adapted to the target motion. 
To reduce the effort of labeling the single frame we offer a semi-automatic bootstrapping technique. To that end, we train a general \soma model with a marker layout containing \numMarkersSuperSet markers selected from the MoSh dataset \cite{Loper:SIGASIA:2014}, visualized in Fig.~\ref{fig:superset_mlayout}; this is a marker super-set.
We choose one sequence per each of representative layouts and run the general \soma to \emph{prime the labels;} we choose one frame per auto-labeled sequence and correct any incorrect labels manually. 
The label priming step significantly reduces the manual effort required for labeling \mocap datasets with diverse marker layouts. After this step, everything stays the same as before.

\textbf{Labeling Active Marker Based \Mocap} should be the easiest case since the markers emit a frequency-modulated light that allows the \mocap system to reliably track them. 
However, often the markers are placed at arbitrary locations on the body so correspondence of the frequency to the location on body is not the same throughout the dataset, hence these archival \mocap datasets cannot be directly solved. %
This issue is further aggravated %
when the marker layout is unknown and changes drastically throughout the dataset. 
It should be noted that, for the case of active marker \mocap systems, such issues could potentially be avoided by a carefully documented capture scenario, yet this is not the case with the majority of the archival data.

As an example, we take DanceDB \cite{DanceDB:2018:aristidou}, a publicly released dance-specific \mocap database. This dataset is recorded by active marker technology from \phasespace \cite{2019phasespace}. 
The database contains a rich repertoire of dance motions with 13 subjects on the last access date. 
We observe a large variation in marker placement especially on the feet and hands, hence we manually label one random frame per each significant variation; in total 8 frames. %
We run the first stage of \mosh independently on each of the selected 8 frames to get a fine-tuned marker layout; a subset is visualized in Fig.~\ref{fig:dancedb_mlyouts}.
It is important to note that we train only one model for the whole dataset while different marker layouts are handled as a source of noise.
As presented in Tab.~6, manual evaluation of %
the solved sequences reveals an above $80\%$ success rate. 
The failures are mainly  due to impurities in the original data, such as excessive occlusions or large marker movement on the body due to several markers coming off (e.g.~the headband).

The second active marker based dataset is Mixamo \cite{mixamoWEB}, which is widely used by the computer vision and graphics community for animating characters. 
We obtained the original unlabeled \mocap marker data used to generate the animations.
We observe more than 50 different marker layouts and placements on the body, of which we pick 19 key variants. 
The automatic label priming technique is greatly helpful for this dataset.

The Mixamo dataset contains many sequences with markers on objects, i.e.\ props, which \soma is not specifically trained to deal with. 
However, we observe stable performance even with challenging scenarios with a guitar close to the body; see the third subject from the left of Fig.~\ref{fig:teaser}.
A large number of solved sequences were rejected mostly due to issues with the raw \mocap data; e.g.\ significant numbers of markers flying off the body mid capture.

\textbf{Labeling Passive Marker Based \Mocap} is a greater challenge for an auto-labeling pipeline. 
For these systems, markers are assigned a new ID on their reappearance from an occlusion, which results in small tracklets instead of full trajectories. 
The assignment of the ID to markers is random.

For the first use case, we process an archived portion of the well-known CMU \mocap dataset \cite{cmuWEB} summing to $116$ minutes of \mocap which has not been processed before, mostly due to cost constraints associated with manual labeling.
It is worth noting that the total amount of available data is roughly 6 hours of which around 2 hours is pure \mpc.
Initial inspection reveals $15$ significant variations in marker layouts, with a minimum 40 markers and a maximum 62; a sample of which can be seen in Fig.~\ref{fig:cmuii_mlyouts}.
Again we train one model for the whole dataset that can handle variations of these marker layouts.
\soma shows stable performance across the dataset even in presence of occasional object props as seen in Fig.~\ref{fig:teaser}; the second subject from the left is carrying a suitcase. 

Due to extreme variation of marker layouts throughout the dataset we notice failure cases where many points could not be assigned to a marker on the body, most probably due to variation from the expected placement. As studied in Sec.~\ref{sec:experiments_mocap_setup_variation}, this deteriorates the labeling performance and could result in a failure in solving the body mainly because of introduced occlusions.

In the second case, we record our own dataset with two subjects for which we pick one random frame and train \soma for the whole dataset. 
In Tab.~\ref{tab:soma_dataset_details} we present details of the dataset motions and per-motion-class performance of \soma.
For this dataset, we manually label it to have ground truth and then we fit the labeled data with \mosh.
This provides ground truth 3D meshes for every \mocap frame.
The V2V error measures the average difference between the vertices of the solved body using the ground truth and using the \soma labels.  
Mean V2V errors are under one $mm$ and usually by an order of magnitude.  
Sub-millimeter accuracy is what users of mocap systems expect and \soma delivers this.

\begin{table}[]
	\scriptsize 
	\centering
	\resizebox{\columnwidth}{!}{%
		\begin{tabular}{ll}
			\hline
			\textbf{Symbol} & \textbf{Description}                     \\ \hline
			\mpc            & \Mocap Point Cloud                       \\
			$\labelSet$       & set of labels including the null label       \\
			$\labelSingle$    & a single label                           \\
			$\labelSetVertexVector$       & vector of marker layout body vertices corresponding to labels not including \nan \\
			$\labelSetVariedVertexVector$ & vector of varied marker layout vertices                                          \\
			$\numMarkers$     & number of markers                        \\
			$\pointSet$       & set of all points                        \\
			$\augasmatGt$   & ground-truth augmented assignment matrix \\
			$\asmatPred$    & predicted assignment matrix              \\
			$\augasmatPred$ & augmented assignment matrix    \\
			$\ascorePred$   & score matrix               \\
			$\lossAsmatWeight$ & class balancing weight matrix                  \\
			\markers        & markers                                  \\
			$\bodyVertex$     & body vertices                             \\
			$\markerDistance$ & marker distance from the body along the surface normal   \\
			$\joints$         & body joints                              \\
			$\numAttHeads$    & number of attention heads                \\
			$\numAttLayers$   & number of attention layers               \\
			&                                          \\ \hline
		\end{tabular}%
	}
	\caption{List of Symbols}
\label{tab:symbols}
\vspace*{-1.5em}
\end{table}

\section{List of Symbols}
In Tab.~\ref{tab:symbols}, we provide a table of mathematical symbols used throughout the paper.

\end{document}